\documentclass[preprint,12pt]{elsarticle}
\usepackage{subfigure}
\usepackage{makeidx}  
\usepackage{multirow}
\usepackage{booktabs}
\usepackage{tabularx}
\usepackage{amsmath,amsfonts,amssymb}
\usepackage{multirow,array}
\usepackage{graphicx}
\usepackage{rotating}
\usepackage{MnSymbol,wasysym}
\usepackage{url,soul}
\usepackage[switch]{lineno}
\usepackage{blindtext}
\usepackage{comment}
\usepackage[flushleft]{threeparttable}
\usepackage{adjustbox}
\usepackage{makecell}
\usepackage{enumitem}
\usepackage{color}

\begin{document}
\begin{frontmatter}
\title{\vspace{-8mm}Measuring the Confidence of Traffic Forecasting Models: Techniques, Experimental Comparison and Guidelines towards Their Actionability}




\author[TECNALIA]{Ibai La\~{n}a}\corref{cor1} \author[TECNALIA]{Ignacio (I\~{n}aki) Olabarrieta} \author[TECNALIA,EHU]{Javier Del Ser}
\address[TECNALIA]{TECNALIA, Basque Research \& Technology Alliance (BRTA), 48160 Derio, Spain}
\address[EHU]{University of the Basque Country (UPV/EHU) 48013 Bilbao, Spain\vspace{-10mm}}
\cortext[cor1]{TECNALIA, P. Tecnologico Bizkaia, Ed. 700, 48160 Derio, Spain. Tl: +34 946 430 50. Fax: +34 901 760 009. E-mail: ibai.lana@tecnalia.com.}

\begin{abstract}
The estimation of the amount of uncertainty featured by predictive machine learning models has acquired a great momentum in recent years. Uncertainty estimation provides the user with augmented information about the model's confidence in its predicted outcome. Despite the inherent utility of this information for the trustworthiness of the user, there is a thin consensus around the different types of uncertainty that one can gauge in machine learning models and the suitability of different techniques that can be used to quantify the uncertainty of a specific model. This subject is mostly non existent within the traffic modeling domain, even though the measurement of the confidence associated to traffic forecasts can favor significantly their actionability in practical traffic management systems. This work aims to cover this lack of research by reviewing different techniques and metrics of uncertainty available in the literature, and by critically discussing how confidence levels computed for traffic forecasting models can be helpful for researchers and practitioners working in this research area. To shed light with empirical evidence, this critical discussion is further informed by experimental results produced by different uncertainty estimation techniques over real traffic data collected in Madrid (Spain), rendering a general overview of the benefits and caveats of every technique, how they can be compared to each other, and how the measured uncertainty decreases depending on the amount, quality and diversity of data used to produce the forecasts.

\end{abstract}
\begin{keyword}
Uncertainty estimation, confidence, traffic forecasting.
\end{keyword}
\end{frontmatter}


\section{Introduction}

Road traffic forecasting has been a subject of intense academic research for decades \cite{lana2018road}. This modeling task was originally approached based on regressive techniques for time series analysis \cite{ahmed1979analysis, levin1980forecasting, moorthy1988short}, veering over time towards Machine Learning (ML) methods \cite{vla7}. When compared to traditional regression techniques, ML allows for the detection and characterization of hidden relationships among data in exchange for a higher level of complexity. A glance at the recent literature contributed on ML-based traffic forecasting reveals that deep neural networks (also referred to as \emph{Deep Learning}) conform nowadays the core of almost any traffic forecasting work \cite{nagy2018survey, manibardo2020transfer}, yielding a torrent of works that are not only abundant, but increasing steadily every year\cite{ermagun2018spatiotemporal, jiang2022graph}. This sustained interest of the research community in this topic suggests that traffic forecasting still poses challenges for the method used to address this modeling task. However, several comprehensive studies expose that traffic forecasting has reached its modeling performance asymptote, wherein the addition of extra layers of complexity in the modeling proposal yields negligible performance gains \cite{manibardo2020deep, yin2020comprehensive}. This prevailing performance-driven pursuit for new traffic forecasting models is far from being actionable from the perspective of Intelligent Transportation Systems \cite{lana2021data}. As a result, most of the challenges of this research area identified years ago remain unsolved to date \cite{lana2018road,vlah14}. 

This noted lack of actionability in current traffic forecasting solutions can be addressed by moving the research focus away from performance metrics towards augmenting their output with valuable information to support practical decision making processes of traffic managers. When the quality of estimated forecasts reaches enough quality in terms of model predictive error, it is of utmost practical interest to explore the individual level of \textit{confidence} associated to each forecast. Model error is a quantitative measure of the performance of the model when producing forecasts for a certain set of input traffic values \cite{majda2018model,anthes2018estimating}. However, both the input data and the modeling technique itself may be subject to different sources of uncertainty \cite{de2007uncertainty}. This means that even models with high performance metrics can yield predictions that are inaccurate beyond the threshold within which they are useful \cite{taleb2009errors}, which propagates to a confidence level of the model in its predicted outcomes. It is not possible to gauge the error of a given prediction until the actual value (\emph{ground truth}) occurs. However, measuring the uncertainty under which the prediction is furnished -- in other words, assigning a confidence level to the prediction -- can be approached at inference time using different strategies. Thus, a manager or service consuming such forecasts would not only be informed about the estimated traffic value for a certain location in space and time, but also the confidence under which the predicted value can be trusted. Besides, from the practitioners point of view, knowing the uncertainty also provides information about when to stop trying to improve a model.

In this regard, many disciplines (especially those where decisions made based on the model's output may entail a risk for the human life) have dealt with the quantification of uncertainty in data-based modeling, so that the provision of \textit{augmented predictions} with confidence estimates has already become a matter of intense research. Within the medical domain, for instance, the confidence of ML models is crucial in model-based clinical diagnosis \cite{hilden1978measurement}, and it has been obtained for diverse purposes \cite{papadopoulos2009confidence, yang2009using, balasubramanian2009support, lambrou2010reliable}. Other fields for which the actionability of forecasts is highly relevant have also considered measuring the confidence of predictions, e.g. in the energy sector \cite{zhou2011short, almeida2015prediction} or in weather and climate modeling \cite{neeven2018conformal, scher2018predicting, lee2019confidence}.

The ITS domain is not unfamiliar to the application of confidence estimation techniques, specially in cases where predictions issued by a model play an essential role in decision making processes. For instance, within the autonomous driving field, vehicles need to sense and anticipate the contextual circumstances in which they operate to determine their behavior in uncertain environment. Researchers working on vehicular perception are highly focused on dealing with uncertainty when estimating the next maneuver \cite{hubmann2018automated, noh2018decision, zhang2020efficient} or the trajectory of other vehicles and pedestrians \cite{huang2019uncertainty,choi2019drogon,li2019coordination}. It is also the case of air traffic management, a field comprising multiple subareas in which decisions must be made fast, and errors can entail important operational consequences and high economical costs. As a result, assessing the uncertainty of forecasts has been addressed in this area for diverse purposes, from aircraft trajectory prediction \cite{casado2016trajectory} to the demand estimation of airport facilities \cite{scala2019tackling, chen2017air} to help managers scale services. In relation to the latter example, and closer to the road traffic domain, a considerable body of literature related to forecasting uncertainty has focused on travel demand estimation. For instance, authors in \cite{parthasarathi2010post, nicolaisen2014ex, yang2013sensitivity} examine the statistical robustness of forecasts and propose different methods to detect them. In \cite{matas2012traffic}, infrastructure capacity constraints are considered to propose a new methodology that quantifies uncertainty, similar to what \cite{hugosson2005quantifying} does for an established and operating travel demand system in Sweden. Analogous research is performed for transport mode choice predictions in \cite {rasouli2014using}, with equivalent actionable results. 

All these research contributions deal with the prediction of travel demand, which is intended to be provided to public authorities, infrastructure and public transportation managers and investors. By informing them about the uncertainty associated to the demand estimation, insights on how traffic and the demand of transportation services will operate in the long term are produced, which can be used to scale appropriately such infrastructures and services. When decision making is based on short-term traffic forecasts, the impact of a forecasting error has a shallower relevance. Anyhow, if these forecasts are meant to be used for any real-world purpose, measuring their uncertainty is of paramount importance for their trustworthiness and actionability. Following up the work in \cite{welde2011planners}, the performance of travel demand forecasts was not properly considered in terms of the uncertainty it was subject to. Up to this point, all the mentioned works involve forecasting up to several minutes or hours. Long-term traffic estimations can also be studied for traffic demand analysis, goal for which measuring their uncertainty becomes an essential actionability driver \cite{lana2019question}. 

This manuscript finds its motivation in the lack of an unified referential work consolidating the state of the art related to uncertainty estimation applied to short-term traffic forecasting problems. To cover this niche, we herein offer a short yet instructive summary about the different types of uncertainty that may arise in road traffic forecasting scenarios, an enumeration of the practical reasons why the confidence of traffic forecasting models is a key driver for their practicality, different uncertainty estimation approaches used nowadays to quantify such a confidence, and scores utilized to compare among such techniques. This multi-faceted analysis is complemented by an extensive experimental benchmark with real traffic and weather data collected over the city of Madrid (Spain). Specifically, the experimental setup and the results obtained therefrom permit to answer with empirical evidence three research questions (RQs) that touch the core of the overall study:
\begin{itemize}[leftmargin=*]
    \item \textit{RQ1} (\textit{Techniques and comparison framework}): How do different uncertainty quantification techniques perform when applied to a particular traffic forecasting model? Which are their main strengths and weaknesses for each problem? Which scores can be used to compare these techniques? What does each of them contribute to the improvement of the model's actionability?
    \item \textit{RQ2} (\textit{Scenario under study}): Which changes in the available data affect uncertainty? How can the confidence of a forecasting model be improved by changing/augmenting data at its input?
    \item \textit{RQ3} (\textit{The relevance of calibration}): What impact does the calibration process of some of the uncertainty estimation techniques have on their outcome? Why is it relevant for traffic forecasting?
\end{itemize}

The rest of this paper is structured as follows: first, Section \ref{sec:uncertainty} introduces the readership to the different sources of uncertainty that one can encounter in traffic forecasting scenarios. This section also enumerates different practical purposes for which the uncertainty of traffic forecasts must be measured, and provides an overview of uncertainty estimation techniques, along with scores that can be used to compare their effectiveness. Section \ref{sec:experiments} introduces the data and the experimental setup devised to answer the RQs formulated above. Section \ref{sec:results} presents and discusses the obtained results, structuring their analysis in terms of the RQ under target. Lastly, insightful concluding remarks are offered in Section \ref{sec:conc}, together with future research directions departing from this study.

\section{Estimating the Confidence of Traffic Forecasting Models: Sources of Uncertainty, Practical Purposes, Techniques and Measures}\label{sec:uncertainty}

As argued in the introduction, uncertainty quantification can improve significantly the worth of any forecasting model. Although it is an old area of study in statistics and probability, its application to machine learning techniques is relatively recent \cite{smith2013uncertainty, sullivan2015introduction}. It has been exposed before that decision makers in many application fields (e.g., medical diagnosis, industrial prognosis) demand the provision of quantitative metrics related to the trustworthiness and reliability of predictions issued by machine learning models. This is certainly not the case of traffic forecasting, a vibrant field with hundreds of publications on a yearly basis, in which confidence measurements are rarely provided. There are some notable exceptions that were surveyed in \cite{de2007uncertainty}, which revealed that most works at the time reporting confidence intervals obtained them experimentally by repeating experiments and measuring the variance and standard deviation of forecasts. Some other works published thereafter took advantage of particular characteristics of Kalman filter modeling techniques to extract confidence intervals for the predictions \cite{guo2010real, guo2014adaptive}. However, this is not a common practice: in the traffic forecasting literature, only performance scores are usually computed. 

Interestingly, machine learning techniques for predictive modeling have reached ever-growing levels of complexity over the years, rendering their modeled knowledge more opaque and difficult to understand for the user consuming their predictions \cite{manibardo2020deep}. As a result, different methods for explaining their inner structure in an user-interpretable manner are under active investigation lately \cite{BARREDOARRIETA202082}. As a supplement to such explainability methods, informing about the uncertainty associated to the predictions supports even further the trustworthiness of forecasting models, and favors decision making processes not only linked to the usefulness of the traffic forecasts themselves (e.g. traffic congestion management), but also in regards to the intelligent collection of traffic data. 

This section delves into these ideas by introducing briefly to the different sources of uncertainty in traffic forecasting (Subsection \ref{ssec:sources}), by discussing on the diverse purposes and inherent usefulness of uncertainty estimation in this context (Subsection \ref{ssec:purposes}), by shortly reviewing state-of-the-art uncertainty estimation techniques (Subsection \ref{ssec:techniques}), and by enumerating different quantitative metrics that can be used to evaluate the quality of estimated confidence intervals (Subsection \ref{ssec:measures}). The overarching goal of this section is to set the knowledge basis towards the experimental part of this work.

\subsection{Sources and Types of Uncertainty in Traffic Forecasting} \label{ssec:sources}

The uncertainty of a model trained with supervised learning can have different kinds of sources. It is commonly accepted that this uncertainty of the output of the model (evaluations of the model with new data) can include a fraction of \textit{epistemic uncertainty}, that accounts for all the uncertainty that the modeling process itself introduces, and \textit{aleatory uncertainty}, that regards the irreducible uncertainty that is present in the data due to their inherent variability \cite{beven2016facets}. However, both the number of different uncertainty sources and their \textit{separability} are open practical questions subject to debate among statisticians, frequently an outset for deeper philosophical discussion \cite{der2009aleatory}. Epistemic uncertainty can (theoretically) be reduced, being it part of the way in which data are structured and represented. On the contrary, aleatory uncertainty cannot be decreased, as it is inherent to the variability and noisy nature of data. Discerning among both sources of uncertainty can be interesting precisely for reducing the share of epistemic uncertainty. Nonetheless, burrowing through the philosophical grounds of the very definition of uncertainty \cite{lindley2000philosophy}, it is unclear whether different sources of uncertainty can be disentangled from each other, or even that the uncertainty associated to data can be isolated from other data features.
\begin{figure}[h!]
    \centering
    \includegraphics[width=0.7\columnwidth]{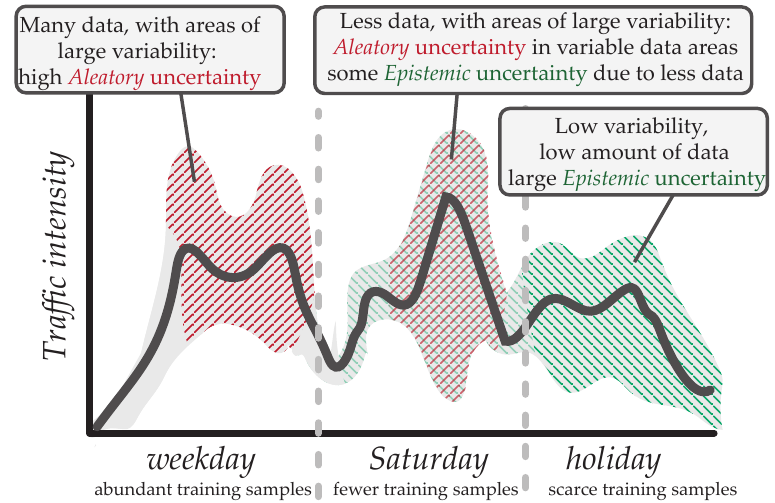}
    \caption{Sources of uncertainty present in traffic forecasting.}
    \label{fig:typesuncert}
\end{figure}

Practical implications of segregating aleatory and epistemic uncertainty in traffic forecasting are clear: while epistemic uncertainty can be reduced to an extent depending on the modeling choice in use, measuring the amount of aleatory uncertainty can dictate whether the addition of new variables and/or the collection of new data helps reducing it effectively. Unfortunately, from a practical point of view such a segregation of uncertainty types might not be feasible \cite{der2009aleatory}. Moreover, it might be even irrelevant as long as mechanisms are designed to reduce the \textit{global} uncertainty that aggregates both sources. In fact, not all uncertainty estimation techniques discussed in this work provide separated uncertainty measurements, consideration that will be contemplated in their experimental comparison.

\subsection{Uncertainty Estimation in Traffic Forecasting: What for?} \label{ssec:purposes}

The use of uncertainty estimation techniques in some of the aforementioned fields is straightforward: quantifying the confidence under which a model produces its output is critical for its actionability in many diverse applications. For this reason, research about uncertainty estimation methods is growing and becoming available for very diverse machine learning models. Short-term traffic forecasting has probably a less critical nature than, for instance, medical or finance sectors. However, measuring uncertainty of traffic forecasts can also be profitable if we consider the purposes for which it can be performed. Such purposes are graphically illustrated in Figure \ref{fig:purposes} and explained in what follows:
\begin{itemize}[leftmargin=*]
    
    \item \textit{Improved actionability}: the unequivocal application of estimating uncertainty is to provide practitioners with a confidence interval of predictions, so that better decisions can be made \cite{lana2021data}. Unlike customary regression scores, which are commonly obtained for a whole test dataset and apprise about the general (averaged) performance of the trained model, uncertainty metrics reach the individual prediction level, giving confidence information for each data point predicted by the model. This not only provides valuable information about the extent to which each individual prediction can be trusted, but also allows for a further study on how traffic behaves in different time intervals and seasons over the year. 
    
    \item \textit{Model selection and comparison}: as explained before, a part of the total uncertainty of the output of a forecasting scheme can be attributed to the way the information is modeled. Uncertainty can be, therefore, another dimension in a model comparison framework, allowing researchers and ITS experts to choose models that perform with narrower confidence intervals. Models performing similarly in terms of regression error but introducing less epistemic uncertainty, can be considered more desirable in some contexts. Furthermore, examining the uncertainty associated to individual predictions can be also interesting to select models that reduce the uncertainty in critical segments of the traffic data series (e.g. in hours featuring high traffic intensity). 
    
    \item \textit{Feature selection}: prior to the application of a training algorithm that fits the parameters of the model to represent a reality represented by data, raw traffic information is arranged into a dataset with a feature space $\mathcal{X}$ and an output space $\mathcal{Y}$. Even before the training process starts, the way in which variables are selected and processed for constructing $\mathcal{X}$ and $\mathcal{Y}$ can affect their degree of uncertainty \cite{der2009aleatory}. This uncertainty should be deemed as epistemic, as it falls under the way knowledge is represented and usually is subject to the experience of the expert at hand. From this conceptual point of view, it is interesting to highlight that the way in which input data are selected, shaped and fed to the model towards its training must be also considered a \emph{part of the model}. Thus, analogously to the previous concept, uncertainty metrics can help, within the framework of a benchmark, to define which input features provide more trustworthy results. Likewise, metrics can be helpful to seek features that help reducing the uncertainty induced by the overall forecasting model. 
    
\begin{figure}[h!]
    \centering
    \includegraphics[width=\columnwidth]{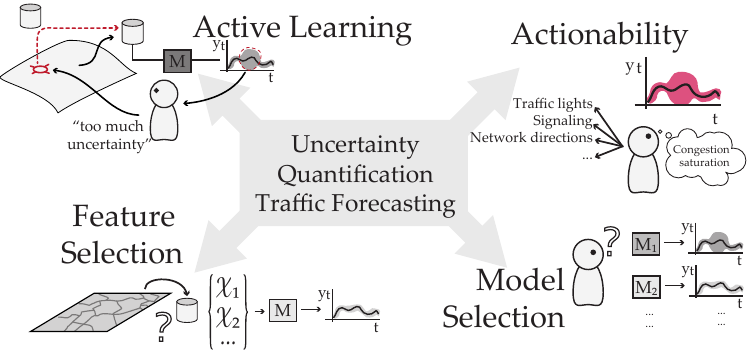}
    \caption{Purposes of uncertainty estimation in the context of traffic forecasting.}
    \label{fig:purposes}
\end{figure}
    
    \item\textit{Active learning:} uncertainty estimation is also an essential component of algorithms that learn actively \cite{shapeev2020active}. Active learning consists of constructing training datasets dynamically, where training samples are added progressively and aimed at reducing uncertainty and increasing diversity. Latest research in this vibrant field is highly reliant on uncertainty estimation \cite{zhu2009active, yang2016active}, and the way in which uncertainty is measured is a matter of study in itself \cite{nguyen2022measure}. In the context of road traffic forecasting, the installation of new traffic reading sensors (either provisional or fixed) can be decided upon the study of the distribution of uncertainty in space and time, especially when traffic measurements are scarce in locations of interest for traffic management. If forecasts for such locations exhibit a higher degree of uncertainty, new measurement campaigns can be commanded to collect more data that allow improving the confidence of the predictions issued by the model. This is especially relevant in practical circumstances where traffic profiles for a certain place are inferred from data collected in other nearby points of the road traffic network \cite{manibardo2022design}.
    
    \item \textit{Trend change detection:} a well-calibrated confidence interval for the output of the model should characterize, with the established confidence level, the percentage of real traffic data that falls within the interval. This \textit{a priori} notion can be used to detect whether, in a traffic data stream, an excessive number of real traffic data points fall outside the confidence interval of the model. If this occurs, it can be an evidence of a change in the underlying distribution of traffic data requiring, upon its detection, an update of the model's knowledge (via incremental training, selective forgetting or any other adaptation method alike). Despite its unquestionable practical benefits (especially in dynamic urban road networks prone to different non-stationarities affecting its traffic flows), this adaptation approach have not been studied until very recently \cite{baier2021detecting}.
\end{itemize}

There are deeper questions related to uncertainty that could concern practitioners, such as the risk of introducing dependence among random events that affect the system being modeled \cite{der2009aleatory}. Nonetheless, the essential actionability aspects presented above can help ITS stakeholders design more confident models, and understand them better in order to put their output to practice.

\subsection{A Brief Overview of Uncertainty Estimation Techniques}\label{ssec:techniques}

Quantifying the uncertainty present in the output of a model implies measuring the way in which it can vary. This can be expressed as the variance or standard deviation of each predicted value, or by reporting different statistics (e.g. confidence intervals or percentiles) of the estimated output distribution \cite{de2007uncertainty}. When dealing with machine learning models, once a dataset is built and the parameters of the chosen model are trained through a training algorithm, the outputs of such model will not vary given a fixed input. For a certain forecasting query, there is only one output given one input to the model. Unless probabilistic formulations of the model's parameters are formulated (as in Bayesian neural networks), most machine learning models are \emph{deterministic} after their parameters have been trained. Thus, with an already trained model, estimating the variability of each forecast requires in general an uncertainty estimation technique. 

On this basis, many techniques and methods have been developed by the community to characterize the output distribution of a machine learning model based on prior knowledge from training data. Once this output distribution is estimated, the usual approach consists of defining significance levels and obtain confidence intervals. Some machine learning models allow changing their default operation to obtain predictions based on percentiles. These model-specific approaches have been described in the literature as \emph{intrinsic} \cite{uq360-june-2021}. Conversely, \emph{extrinsic} uncertainty estimation techniques are not specific to a particular machine learning model, but allow obtaining a measurement of uncertainty once a model has been trained (i.e., a post-hoc estimation), normally through a calibration process. 

Even though some authors have tried to categorize the landscape of uncertainty measuring techniques \cite{uq360-june-2021}, it appears to be difficult to find consensus beyond the differences between intrinsic and extrinsic techniques exposed above. For instance, some techniques allege to be able to discriminate between the epistemic and aleatory uncertainty present in the output of machine learning models \cite{hullermeier2021aleatoric, senge2014reliable}, frequently based on Bayesian formulations of their underlying training mechanisms. Other branch of techniques based on wrapper methods rely on a calibration process prior to the training phase of the algorithm, whereas other techniques hinge on the use of ensembles to approximate the output distribution. Lastly, different variants of deep neural networks have been proposed in the last years to estimate the output distribution, via the probabilistic definition of their trainable weights, evidential formulations of the loss function used to guide the training process, or the derivation of mechanisms to sample the sought distribution. 

Beyond the discussion on the suitability of one taxonomic criterion or another, in what follows we examine the most widely adopted options to estimate the uncertainty in machine learning models used for regression tasks. To this end we pause briefly at the specific techniques organized in the taxonomy shown in Figure \ref{fig:taxonomy}. This short review connects tightly with the experiments designed to answer RQ1, where a comprehensive comparison benchmark over real traffic data will expose the strengths, weaknesses and the applicability of the methods reviewed below:
\begin{itemize}[leftmargin=*]
    \item \textit{Conformal prediction}: as defined in \cite{gammerman2013learning, vovk2005algorithmic}, conformal predictors are \emph{confidence} predictors. Their operation is based on the conception of \emph{nonconformity} measures, which allow predicting regions of points instead of individual points (i.e., confidence intervals) by using the statistical knowledge that can be obtained through a calibration process from the training samples. This method has been prevalent in the machine learning community for years \cite{balasubramanian2014conformal}, being used in applications arising in many fields \cite{johansson2014regression}. Conformal prediction is agnostic to the particularities of the underlying model, and can be used, in principle, with any machine learning algorithm and dataset configuration. 
    
    \item \textit{Ensemble methods}: ensemble learning consists of the aggregation of knowledge obtained from different models learned from different \emph{views} of the data. Combining the predictions issued by different methods learned from the same data has been proven to yield better performance than any contributing member to the ensemble, which is achieved by reducing the spread in the predictions made by such models \cite{dietterich2002ensemble,gonzalez2020practical}. Besides the increased robustness favored by this smoothing process, ensembles represent an easy and straightforward approach to sample the distribution associated to the combined output; once trained, each base learner yields a different prediction for a given input, all of which are aggregated by averaging or other more sophisticated combination procedures (e.g. weighting based on out-of-bag performance estimates). Thus, confidence intervals can be obtained by characterizing statistically the distribution of all the different predictions produced by each learner in the ensemble for a given input \cite{mentch2016quantifying}.
\begin{figure}[h!]
    \centering
    \includegraphics[width=0.8\columnwidth]{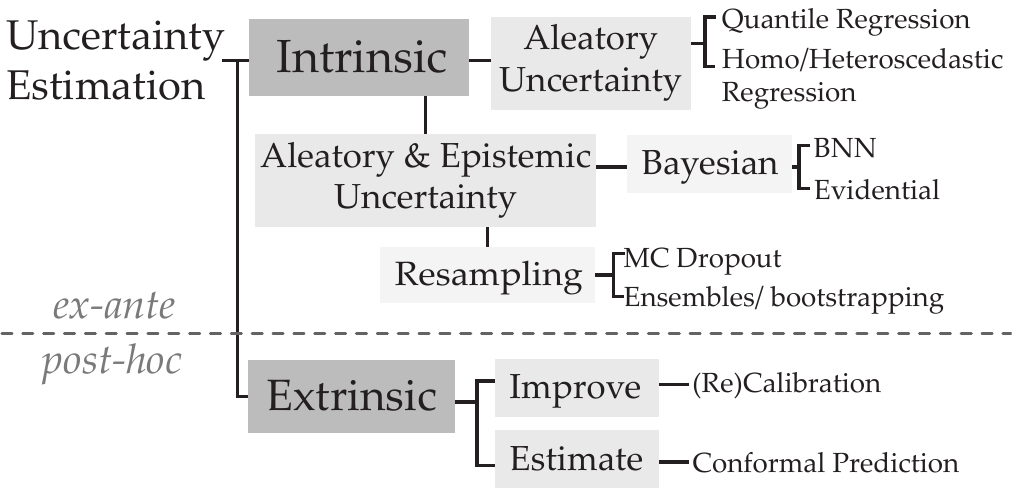}
    \caption{Taxonomy of uncertainty estimation techniques tackled in this work.}
    \label{fig:taxonomy}
\end{figure}

    \item \textit{Quantile regression}: This family of methods was originally conceived for linear regression methods, for which, instead of using the median as an output, asymmetric errors were considered in order to have upper and lower boundaries of an interval defined by a quantile \cite{koenker2001quantile}. More modern contraptions have used this notion as a base to estimate uncertainty of other machine learning models \cite{rahmati2019predicting, vaysse2017using}. Quantiles have been introduced in loss functions of several learning algorithms, from gradient boosting regression to deep neural networks. These reformulation of the loss functions permit to predict not the median, but instead the extremes of an interval defined by the quantile sought. Their main difference with respect to previous methods in terms of prediction interval is the distribution of points inside the interval. Other methods generate predictions that may have an unknown distribution. Once they are characterized, the boundaries of the interval are obtained according to a significance level. This approach \emph{produces only boundary lines}, and they are normally distributed. A characteristic prediction would be the median, but any other \emph{predicted} points inside the interval would be placed according to their significance level. This becomes a relevant issue when comparing the output of these methods to other uncertainty estimation approaches, as the experimental discussions later held will clearly show.
    
    \item \textit{Bayesian inference based algorithms}: in general, these algorithms aim to characterize the posterior probability distribution of the model parameters based on the available data and the assumption of an a priori distribution of the parameters. Once this distribution is characterized after training, we have the possibility of establishing confidence levels. Particularly with deep Bayesian networks, the parameters of the model (weights) are driven by a priori probability distribution (often set to be Gaussian), making the so-called family of Bayesian neural networks a model choice particularly suited to estimate uncertainty in modeling problems \cite{wang2016towards}. Many authors in different fields have taken advantage of Bayesian formulations. For instance, a thorough review of different applications of Bayesian learning approaches in the health domain can be found in \cite{abdullah2022review}, whereas use cases have been reported for other diverse scenarios such as plant disease detection \cite{hernandez2020uncertainty} or solar power forecasting \cite{lee2018bayesian}, among others. More recently, some attempts at applying Deep Bayesian neural networks for the delivery of confidence-aware road traffic forecasts have been done \cite{9533457,9140389}, but efforts in this regard to date have been significantly lower than in other application scenarios.
    
    \item \textit{Model-specific methods}: finally, in this last category we gather all techniques that are specific to a certain family of forecasting models. Many of them have been proposed for deep neural networks, leveraging the flexibility granted by the use of a loss function to guide the training process, or the existence of neural computation mechanisms that allow for an approximate representation of the sought output distribution. A representative technique of the former approach is \textit{evidential deep learning} \cite{amini2020deep}, which was proposed to overcome the high computational complexity of the training process of Bayesian neural networks and the dependence of the estimated uncertainty on the suitability of the assumed priors for the network weights. In doing so, evidential deep learning proposes a framework in which a prior distribution is placed on the statistics of a Gaussian output distribution (as opposed to Bayesian neural networks, where priors are imposed to the network weights), so that the addition of newly observed traffic samples provides more support for the neural network to learn the parameters of the evidential distribution. On the other hand, Montecarlo Dropout \cite{gal2016dropout} resounded loudly in the community for its simplicity and scalability to integrate the model's output likelihood by randomly switching off neurons in a neural network at inference time. Lastly, an \textit{intrinsic} approach included in IBM's UQ360 library \cite{uq360-june-2021} is \textit{heteroscedastic regression}, which takes advantage of the the expected noise of the data in the model to capture both kinds of uncertainty. An alternative is \textit{homoscedastic regression}, which assumes that noise is constant across data points.
\end{itemize}

\subsection{Measuring and Comparing Uncertainty Estimations}\label{ssec:measures}

The output of any of the methods presented above should be regarded as a measure of the amount of uncertainty associated to their predictions. This is not a trivial aspect: as argued in Subsection \ref{ssec:sources}, boundaries between different types of uncertainty are not clear, and not all methods measure  the model's confidence by following the same principles. For this reason, there is a series of metrics that, departing from a common ground, not only provide a notion of the amount of uncertainty, but also indicate whether the measurement of uncertainty itself is performed equally. As in any other regression task, in traffic forecasting the measurement of the uncertainty at the output of a model can be performed by establishing a significance level beforehand (usually denoted as $\alpha$), and by defining confidence intervals that contain a fraction of the possible outputs that matches the defined $\alpha$. Thus, uncertainty relates to the \emph{width} of the interval for each estimated point. For a certain significance level $\alpha$, a narrower interval represents a less uncertain output of the model for which it is estimated. 

Beyond these baseline principles, there is no consensus in the literature regarding these metrics, to the point of referring to the same metric with different names, or using metrics that cannot be measured for all uncertainty estimation methods. With a practical stance, we now summarize a list of  uncertainty metrics that can be used to measure different aspects of the estimated uncertainty, their different usages and practical implications. The adopted nomenclature for the rest of the paper is the one used in \cite{rodrigues2018heteroscedastic}:
\begin{itemize}[leftmargin=*]

\item \textit{Interval width}: this first metric refers to the amplitude between the lowest and highest points of the estimated interval around each predicted data point. The concept of \textit{efficiency} or \textit{informational efficiency} of a conformal predictor \cite{vovk2005algorithmic} is related to this width, while a common uncertainty quantification metric known as \textit{sharpness} \cite{tran2020methods}, which is computed based on the variance at each point, measures essentially the same. The name \textit{informational efficiency} used by some sources in the related literature refers to the way in which the interval informs: a very narrow interval with high confidence is very informative, while a very wide interval that covers a great number of possible values gives less information with the same confidence, thus it is less efficient. This metric can be also found referred to as \emph{Mean Interval Length} (\texttt{MIL}, as in \cite{rodrigues2018heteroscedastic, petersen2021short}) or \emph{Mean Prediction Interval Width} (\texttt{MPIW},  \cite{uq360-june-2021, shrestha2006machine}). Given a significance level $\alpha$, this metric expresses how wide must the interval be to include the fraction of values defined by $\alpha$. As the width can be different for each traffic forecast, it is common to obtain an averaged value of \texttt{MIL} (or \texttt{MPIW}). Mathematically, this metric is defined as:
\begin{equation}\label{eq1}
\texttt{MIL}=\frac{1}{T}\sum_{t=1}^{T} (u_{t} - l_{t}),
\end{equation}
where $u_t$ and $l_t$ are the upper and lower boundaries of the confidence interval estimated by the technique at hand, and $T$ is the number of instances in the test dataset. 
                    
\item \textit{Interval coverage}: this second metric is the fraction of the possible outputs that is covered by the interval. This can be related to the so-called \textit{validity} metric \cite{vovk2005algorithmic}, and has been also referred to as \emph{Interval Coverage Percentage} (\texttt{ICP},  \cite{rodrigues2018heteroscedastic, petersen2021short}) or \emph{Prediction Interval Coverage Probability} (\texttt{PICP}, \cite{uq360-june-2021, shrestha2006machine}). This metric provides a notion of how valid the interval is, as its value should concur with the percentage of samples defined by the established significance level ($1-\alpha$). The metric is defined as per eq. \ref{eq2}:
\begin{equation}\label{eq2}
\texttt{ICP}=\frac{1}{T} \sum_{t=1}^{T} \mathbb{I}(l_{t} \leq y_t \leq u_{t}),
\end{equation}
where $l_t$ and $u_t$ are the lower and upper confidence interval boundaries as in Expression \eqref{eq1}, $y_t$ denotes the ground truth value associated to the $t$-th predicted value $\widehat{y}_t$ in the test set, and $\mathbb{I}(\cdot)$ denotes an auxiliary binary function taking value $1$ if its argument is true ($0$ otherwise).

\item \textit{Interval width with relation to the forecasting error}: \texttt{RMIL} \cite{rodrigues2018heteroscedastic} is a variant of \texttt{MIL} that relates the size of the estimated confidence interval to the error of the forecast. This allows for larger intervals in order to cover those cases with largest forecasting error (i.e., those traffic values that result to be more difficult to forecast precisely). As a result, \texttt{RMIL} permits to compare two forecasting models that deal with different input predictors in terms of uncertainty. The metric is defined as:
\begin{equation}\label{eq3}
\texttt{RMIL}=\frac{1}{T}\sum_{t=1}^{T}\frac{(u_{t} - l_{t})}{\left|y_t-\widehat{y}_t\right|},
\end{equation}
where $y_t$ is the real value for the $t$-th test instance, and $\widehat{y}_t$ denotes the predicted value for that instance issued by the forecasting model. This metric can return very high (even infinite) values if the prediction is very close to the real values, reason for which it has not been considered for the experimentation of this work. 

\item \textit{Calibration curves}: The calibration of a forecasting model aims to achieve a statistical consistency between the distribution of the forecasts and the distribution of the real values \cite{gneiting2007probabilistic}. The outputs of a well-calibrated forecasting model will follow a similar distribution to the real values of the ground truth the model is representing. This property is explored through calibration curves, i.e., a two-dimensional plot relating every predicted value to the real observation it approximates \cite{uq360-june-2021}: as such, the optimal calibration curve is the identity function, whereas the deviation of the curve with respect to the optimal calibration is denoted as \textit{calibration error}. It is relevant to note that, although previous metrics can be comparable for any uncertainty estimation approach, the way in which output distributions are obtained may affect substantially to the measurement of the calibration curve and error. 

\item \textit{Other metrics}: Gneitig et al. \cite{gneiting2007strictly} proposed a set of \textit{proper scoring rules} that cover, among others, diverse probabilistic aspects of interval forecasts, including combined metrics of some of the aforementioned scores, e.g., \emph{calibration} and \emph{sharpness}. Metrics like Negative Log Likelihood, Interval Score or Check have been used in a variety of uncertainty estimation works \cite{carvalho2016overview},  providing a different perspective to the analysis of the size of the intervals and their coverage of real samples. Considering that they provide similar insights as the above metrics, and there are some hindrances to apply them to all of the uncertainty estimation methods (for instance, some of them are only intended for quantile-based methods), we have prioritized the metrics introduced previously in our experiments for the sake of a clearer and more uniform analysis. 
\end{itemize}

These metrics can be used to assess the impact of different datasets and configurations of the forecasting model, as well as to examine the convenience of each of the uncertainty quantification methods. The insights extracted from their analysis will be helpful answering the different research questions of the experimental study discussed in the next section.

\section{Experimental Setup}\label{sec:experiments}

In order to assess and compare the uncertainty estimation techniques and confidence metrics presented above, and to discuss on their implications for the actionability of traffic forecasts, an experimental setup has been defined, which comprises several scenarios with real-world traffic data and different features. To this end, the setup relies on the extensive collection of traffic flow readings made public by the Madrid City Council in its Open Data portal (\url{https://datos.madrid.es/}), which releases a 5-year historic record of 15-minute traffic flow observations collected in more than 3,800 locations over this city. With these data, together with meteorological and calendar information, different datasets $\mathcal{X}$ have been composed. Hence, an scenario is defined based on the datasets in use, the considered forecasting horizon (i.e., the gap between the query time and the time at which the target variable sought occurs), and the forecasting model used to model the relationship between its input variables and the traffic measure to be predicted. By examining the relative behavior of the confidence metrics in each one of such scenarios, we can 1) gauge how each dimension of the scenario affects uncertainty (in response to RQ2), and 2) compare uncertainty estimation methods in terms of the metrics in different contexts (providing informed insights for RQ1).

\subsection{Datasets}

\begin{figure}[h!]
    \centering
    \includegraphics[width=0.85\columnwidth]{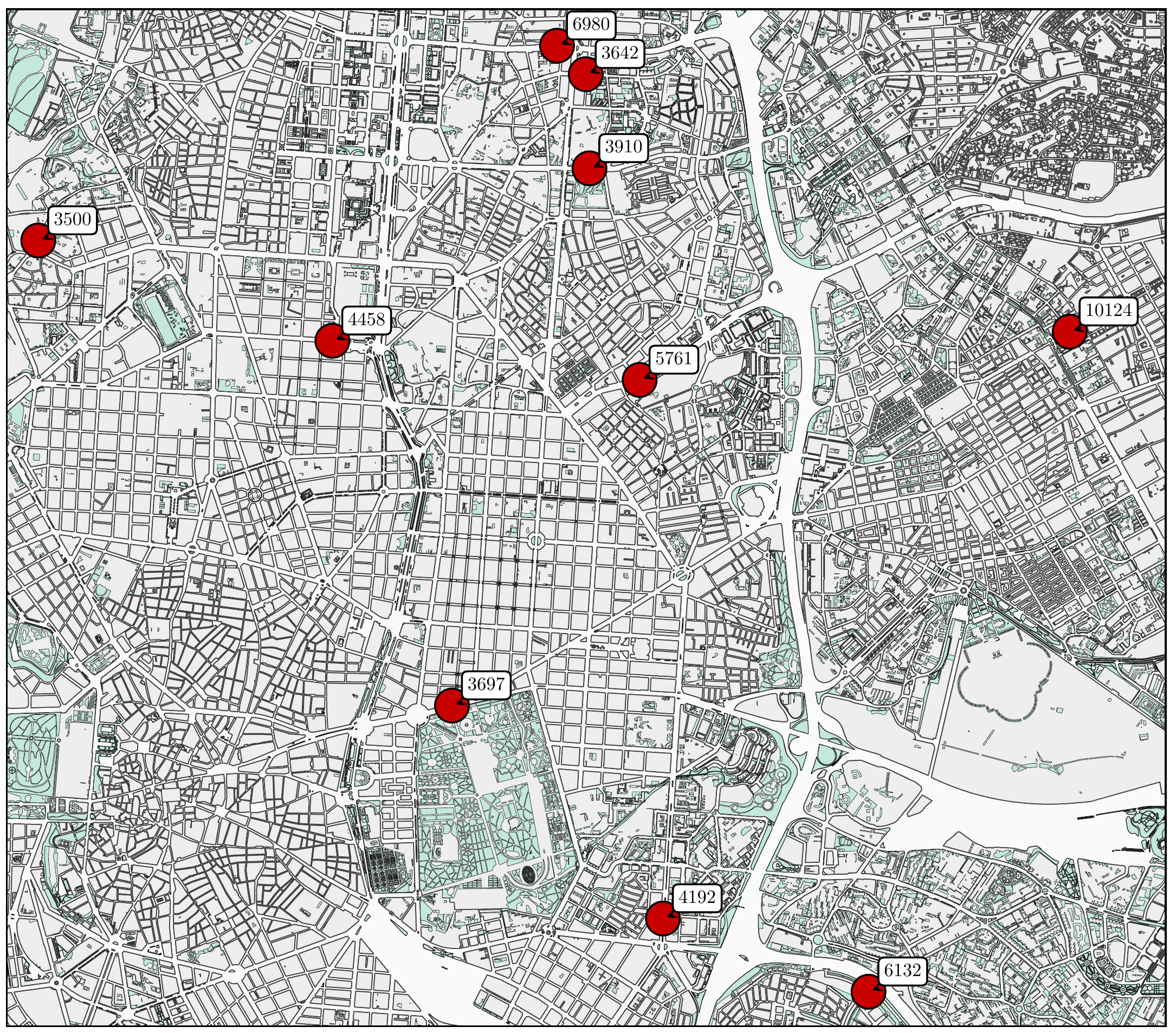}
    \caption{Distribution of the selected traffic loops located in the city center of Madrid (Spain), whose collected data have been used for the experiments reported in this work.}
    \label{fig:map}
\end{figure}

The data used for defining the scenarios introduced above are based on real urban traffic data collected in 10 different locations of the city of Madrid (Spain). The selection of inductive loops has been made according to the variability of the surrounding urban topology of every sensor, in order to analyze the potential relationship between the traffic profiles in those locations and the measured uncertainty of a modeling pipeline aimed to forecasts those traffic measures. The spatial distribution of the selected sensors, depicted in Figure \ref{fig:map}, shows that some loops are placed in locations with heavy traffic conditions (i.e., references\footnote{These numerical references correspond to the labels assigned to the loops in the Open Data portal from which data were retrieved.} 3697, 3910 and 5761), while others are placed in small roads of residential areas or main arterial roads traversing commercial districts.

\begin{figure}[h!]
    \centering
    \includegraphics[width=0.95\columnwidth]{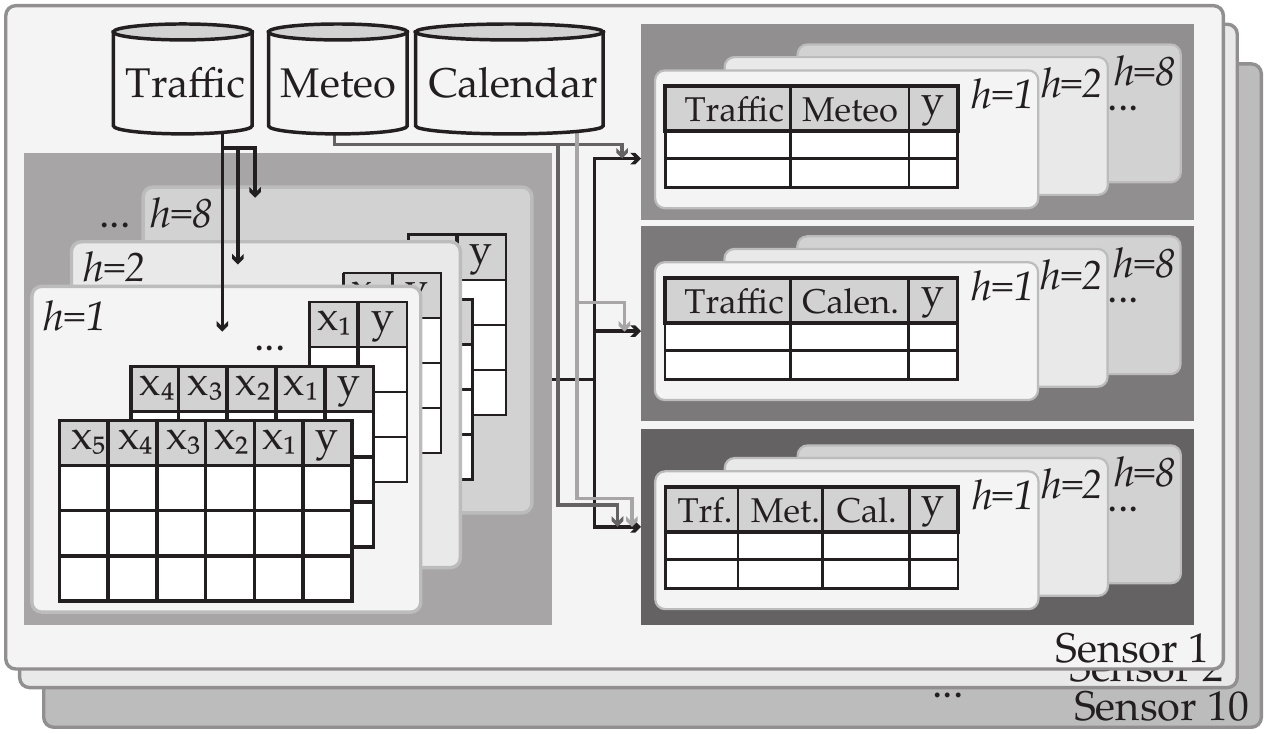}
    \caption{Datasets created for the experiments.}
    \label{fig:data}
\end{figure}

Data collected for the whole year 2019 were built into datasets to train and test different forecasting models. Such datasets include information of up to 5 steps (i.e., 1 hour and 15 minutes) of past traffic measurements before the traffic point to be predicted ($t_{-4}$, $t_{-3}$, $t_{-2}$, $t_{-1}$, and $t_{0}$). This window of observations will be configured as the input to the models in 2 forms $\omega=\{1,5\}$: either all past measurements ($\omega=5$) or just one ($\omega=1$). For each case, different forecasting horizons $h = \{1, 2, 4, 8\}$ are considered, producing predictions in instants $t_{+1}$, $t_{+2}$, $t_{+4}$ and $t_{+8}$, and providing traffic forecasts of up to two hours in the future. Datasets may include (or not) meteorological information (temperature, cloud cover, humidity and precipitation intensity) and calendar information regarding local and national holidays, academic and scholar calendar and day of the week. The availability of this data at the input to the models is denoted by two binary variables $m=\{0,1\}$ (meteorology) and $c=\{0,1\}$ (calendar). With these different types of input variables, datasets with all possible combinations are created following the scheme in Figure \ref{fig:data}, leading to a collection of 320 datasets resulting from $10$ loops $\times$ $2$ window lengths $|\omega|$ $\times$ $|m|$ (meteorological information available or not) $\times$ $|c|$ (calendar information available or not) $\times$ $4$ forecasting horizon values. Models will be tested and their uncertainty measured with different sets of these combinations, in order to extract insights and answer each of the research questions stated in the introduction.

Additionally, a standardization procedure was applied to all data to guarantee the correct operation of some of the models, particularly those that are sensitive to differences in the statistical support of their predictors. Lastly, datasets were split into train and test partitions, stratifying them across all months of the year available for training. For each month, 3 weeks were considered as train data, whereas 1 week was left for test data. This permits to include all kinds of traffic profiles (which are highly variable throughout the year) in the training dataset, and also allows for a proper train-calibration-test split for the Conformal Prediction approach: the calibration set will consist of the $3^{rd}$ week of each month, thus providing data from all along the year to calibrate the models. Techniques that do not include a calibration step do not use this part of the dataset as training data, so all the models receive the same training information. 

\subsection{Methodological Approach}

\begin{figure}[h!]
    \centering
    \includegraphics[width=0.9\columnwidth]{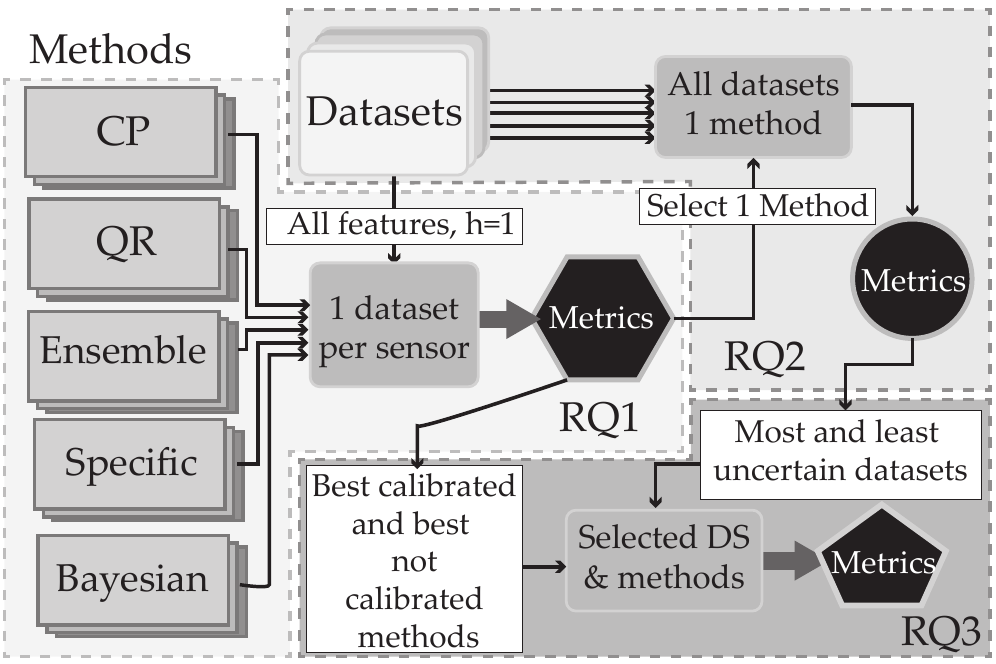}
    \caption{Schematic diagram of the processing pipeline applied to the traffic forecasting datasets under analysis.}
    \label{fig:pipeline}
\end{figure}

Departing from the collection of 320 datasets defined in the previous section, a processing pipeline is established. The steps are schematically depicted in Figure \ref{fig:pipeline}. 

To answer RQ1, several predictive methods that cover all the uncertainty estimation approaches described in Section \ref{ssec:techniques} are considered. Prediction tests are conducted with each model and the dataset combination comprising all features ($m=c=1$) and the forecasting horizon set to $h=1$. The uncertainty of each (model, dataset) combination is then estimated with the available techniques. A relevant remark related to this first comparison among techniques is that not all techniques are suitable for all algorithms. For instance, measuring uncertainty through techniques closely linked to specific elements of deep neural networks (e.g., Monte Carlo dropout) are not available for tree-based shallow learning algorithms. Thus, the comparison of techniques will be influenced by the capacities and limitations of the underlying predictive models, as well as by the \textit{epistemic uncertainty} they introduce. Table \ref{table:algorithms} shows the specific models that have been considered for each uncertainty estimation technique in our experimental benchmark.
\begin{table}[ht]
\centering
        \caption{Predictive models used in the case study and the uncertainty techniques applicable to each of them}
        \label{table:algorithms}
        \vspace{3mm}
        \resizebox{1\columnwidth}{!}{\begin{tabular}{ccccccl} 
         \toprule
         \multirow{1}{*}{\textbf{Model}}  &
         \multicolumn{1}{c}{\makecell{Conformal\\Prediction}} &
         \multicolumn{1}{c}{Ensemble}  &
         \multicolumn{1}{c}{\makecell{Quantile\\Regression}} &
         \multicolumn{1}{c}{Bayesian} &
         \multicolumn{1}{c}{Specific}  &
         \multicolumn{1}{c}{Details}   \\
         \cmidrule(lr){1-7}
         \textbf{RFR} & CP-RFR & E-RFR & $\times$ & $\times$ & $\times$& {100 estimators}\\
         \textbf{ETR} & CP-ETR & E-ETR & $\times$ & $\times$ & $\times$ & {100 estimators}\\
         \textbf{GBR} & CP-GBR & $\times$ & Q-GBR & $\times$ & $\times$ & {loss: quantile, 100 estimators}\\
         \textbf{ABR} & CP-ABR & E-ABR & $\times$ & $\times$ & $\times$ & {100 estimators, learning rate: 0.1}\\
         \textbf{MLP} & CP-MLP & $\times$& $\times$ & BNN & $\times$ & Dense(50), 20 epochs, Adam \\
         \textbf{DL} & $\times$ & E-DL & Q-DL & $\times$ & EvDL, MCD, HR & \makecell[tl]{LSTM(50, ReLu), Dense(20, ReLu), 20 epochs, Adam} \\
         \bottomrule
         \multicolumn{7}{l}{}\\
         \multicolumn{7}{l}{RFR: Random Forest Regressor; ETR: Extra Trees Regressor; GBR: Gradient Boosting Regressor; ABR: Ada Boost Regressor;}\\
         \multicolumn{7}{l}{MLP: Multilayer Perceptron;DL: Deep Learning; DL-Q: Deep Learning with Quantile Loss; BNN: Bayesian  Neural Network;}\\
         \multicolumn{7}{l}{HR: Heteroscedastic Regression; EvDL: Evidential Deep Learning; MCD: MonteCarlo Dropout}
        \end{tabular}}
\end{table}

In addition to the predictive performance of the considered forecasting models (which may differ among them), the comparison benchmark also accounts with the confidence metrics described in Section \ref{ssec:measures}. By simultaneously reporting on both confidence and forecasting error, we will be able to compare them in model-agnostic dimensions, connecting clearly with the responses sought to answer RQ1. When it comes to RQ2, a single uncertainty estimation method and predictive model will be selected, based on the results obtained from the prior experimentation done in regards to RQ1. The rest of dataset combinations (different horizons and different sets of features) will be used to obtain forecasts, estimate uncertainty and assess their impact on the confidence levels. Lastly, tests with calibration-based and non-calibrated techniques will be conducted using different combinations of datasets, so that RQ3 can be replied in an empirically informed fashion.

\section{Results and Discussion} \label{sec:results}

\begin{figure}[h!]
    \centering
    \includegraphics[width=0.85\columnwidth]{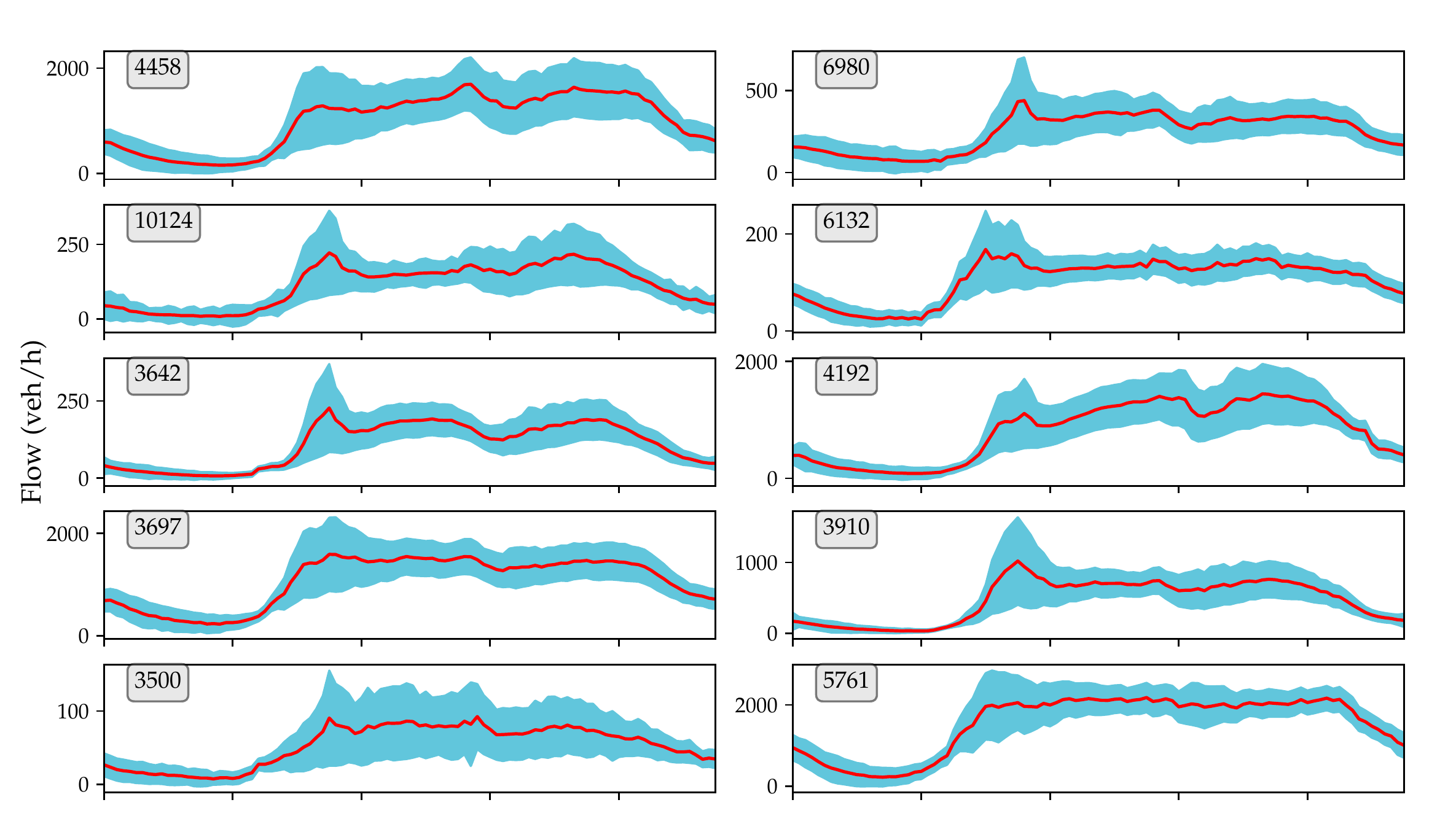}
    \caption{Average flow in each location (red line). The shaded area represents $\pm$ one standard deviation.}
    \label{fig:trafficdata}
\end{figure}

We now discuss on the experimental results of the setup described in Figure \ref{fig:pipeline}. In doing so, we organize our examination of such results in terms of the research questions posed in the introduction. Models and uncertainty estimation techniques have been tested with different sensors placed in around the city in order to have different traffic profiles and also assess the impact of the traffic behavior on the modeling tools. Thus, traffic profiles of the $10$ traffic sensors are shown on Figure \ref{fig:trafficdata}. Each plot nested in this figure represents the average traffic (computed over 96 daily traffic traces) of every sensor in the $i^{th}$ 15-minute interval of a day, where the standard deviation is also provided as a shaded area to illustrate the intra-day variability of the traffic profiles at each sensor's location. 

\begin{figure}[h!]
    \centering
    \includegraphics[width=0.9\columnwidth]{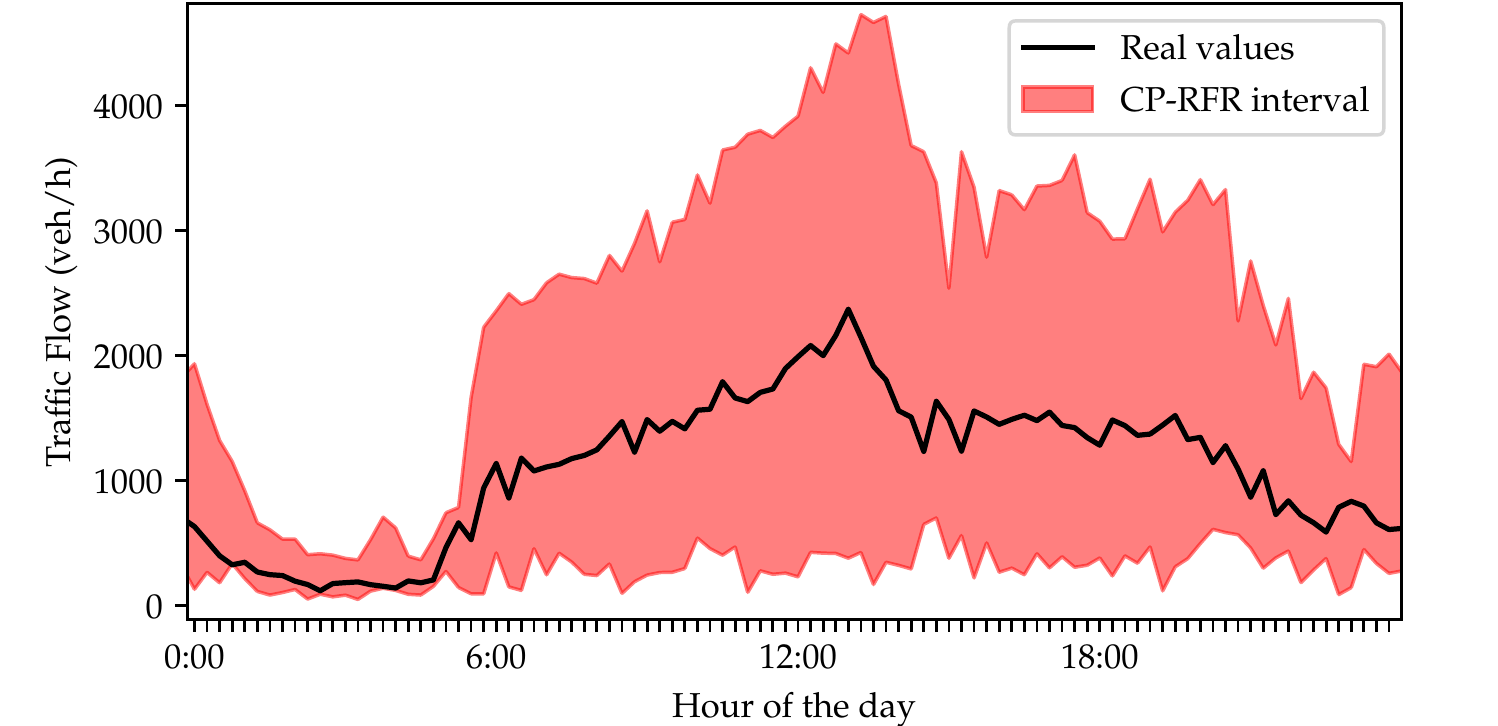}
    \caption{One day of traffic data collected in loop 4458. Real values are shown on top of the confidence intervals estimated by conformal prediction for a Random Forest forecasting model.}
    \label{fig:intervals}
\end{figure}

As revealed in this figure, sensors such as $3642$, $6132$ or $3500$ have very low traffic profiles, with less than 300 vehicles/hour on average for the busiest hour. This implies more real traffic data close to 0 vehicles/hour and larger relative errors in the prediction. Large deviation areas like the one in sensor $4458$ anticipate larger confidence intervals, while loops like $4192$ with narrower deviations and wider dynamic range are expected to provide better performing and less uncertain models. As a sample of how intervals are produced, a single day of data collected by sensor $4458$ is shown on Figure \ref{fig:intervals}, in which the real data and the confidence intervals produced by the combination of conformal prediction (CP) and a Random Forest regressor (RFR) are presented. The interval is highly informative, as its width unveils when the most uncertain predictions are produced, and how uncertain they may be. A prediction at noon with a $90\%$ confidence will be quite uncertain. 

Once the inherent variability of traffic data has been shown, we proceed by discussing on the quantitative results produced for addressing every research question:

\subsection{RQ1: How do different uncertainty quantification techniques perform when applied to a particular traffic forecasting model?}

The first research question regards the performance of the uncertainty estimation techniques considered in our benchmark. As indicated in Table \ref{table:algorithms}, not all of such techniques operate in the same fashion, reason for which the metrics proposed in Section \ref{ssec:measures} are used as a standardized framework that allows comparing them under unified criteria. 

\begin{figure}[hb!]
    \centering
    \includegraphics[width=\columnwidth]{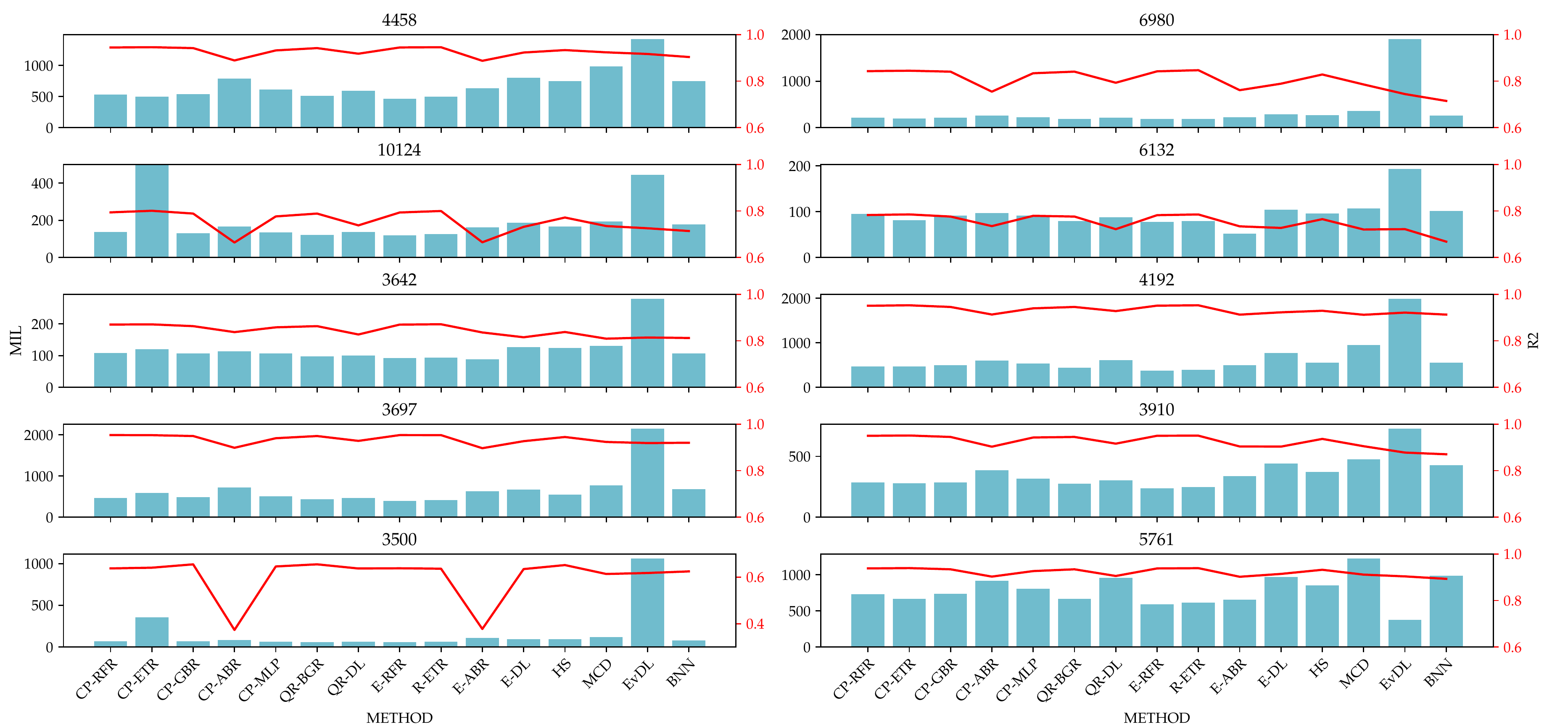}
    \caption{Mean Interval Length (\texttt{MIL}) obtained by the uncertainty estimation methods over different forecasting models learned from the data collected by every loop. The forecasting horizon is $h=1$. Predictive performance is presented in red ($R^2$ score).}
    \label{fig:MIL}
\end{figure}

We start our discussion around RQ1 by inspecting Figure \ref{fig:MIL}, which shows the mean interval length, or the average width of the interval (\texttt{MIL}) obtained for the datasets comprising meteorological and calendar features ($m=c=1$), using each predictive model and uncertainty estimation approach. On top of each bar plot, the performance score $R^2$ is shown with a red line. As expected, performance is generally lower for those sensors with lower traffic profiles, particularly for $3500$, which scores the worst with $R^2$ values not surpassing beyond $0.7$. The lower performance of ADABoost based methods (namely, CP-ABR, E-ABR) is specially visible here, but can be also noticed in the rest of sensors. This predictive method suffers the burden of removing data or removing estimators for estimating uncertainty, and in general is not a suitable option if confidence is to be measured. On the other hand, deep learning approaches (including heteroscedastic regression) present generally wider intervals than conformal prediction approaches, with a noteworthy increase when using evidential formulations of deep learning models. Indeed, the confidence intervals provided by evidential Deep Learning are significantly wider than those elicited by other uncertainty estimation techniques, except for sensor $5761$. This can be explained by the renowned sensitivity of evidential formulations with respect to its regularization
coefficient $\lambda$ \cite{amini2020deep}. This value was left equal to a constant value ($1.0$) for all experiments so as to expose this known issue. 

On the other hand, conformal prediction (CP) based methods perform very similarly to each other across different traffic loops in terms of the \texttt{MIL} metric. This holds in all cases except for CP-ETR in two of the sensors ($10124$ and $3500$). The removal of certain nodes that does not happen in the very similar Random Forest model produces extreme peaks in the estimation of the interval boundaries when real observations are very close or equal to 0. This explains these high values that are produced after averaging these peaks with the rest of the points of the boundary, which are essentially almost the same as for Random Forest. Conformal prediction, quantile regression or ensemble methods produce intervals of very similar widths for most of the cases, while Bayesian methods present an irregular behavior in terms of interval width and a subpar predictive performance reflected in generally lower $R^2$ values.

Measuring the average interval width through the lens of the \texttt{MIL} metric shows one face of the uncertainty associated to a traffic forecast: the narrower the interval, the more trustworthy the prediction can be thought to be. However, too narrow intervals can leave too many real samples outside them. This reason motivates the adoption of \texttt{ICP} as the second confidence metric for the discussion of this first set of results. 
\begin{figure}[h!]
    \centering
    \includegraphics[width=0.95\columnwidth]{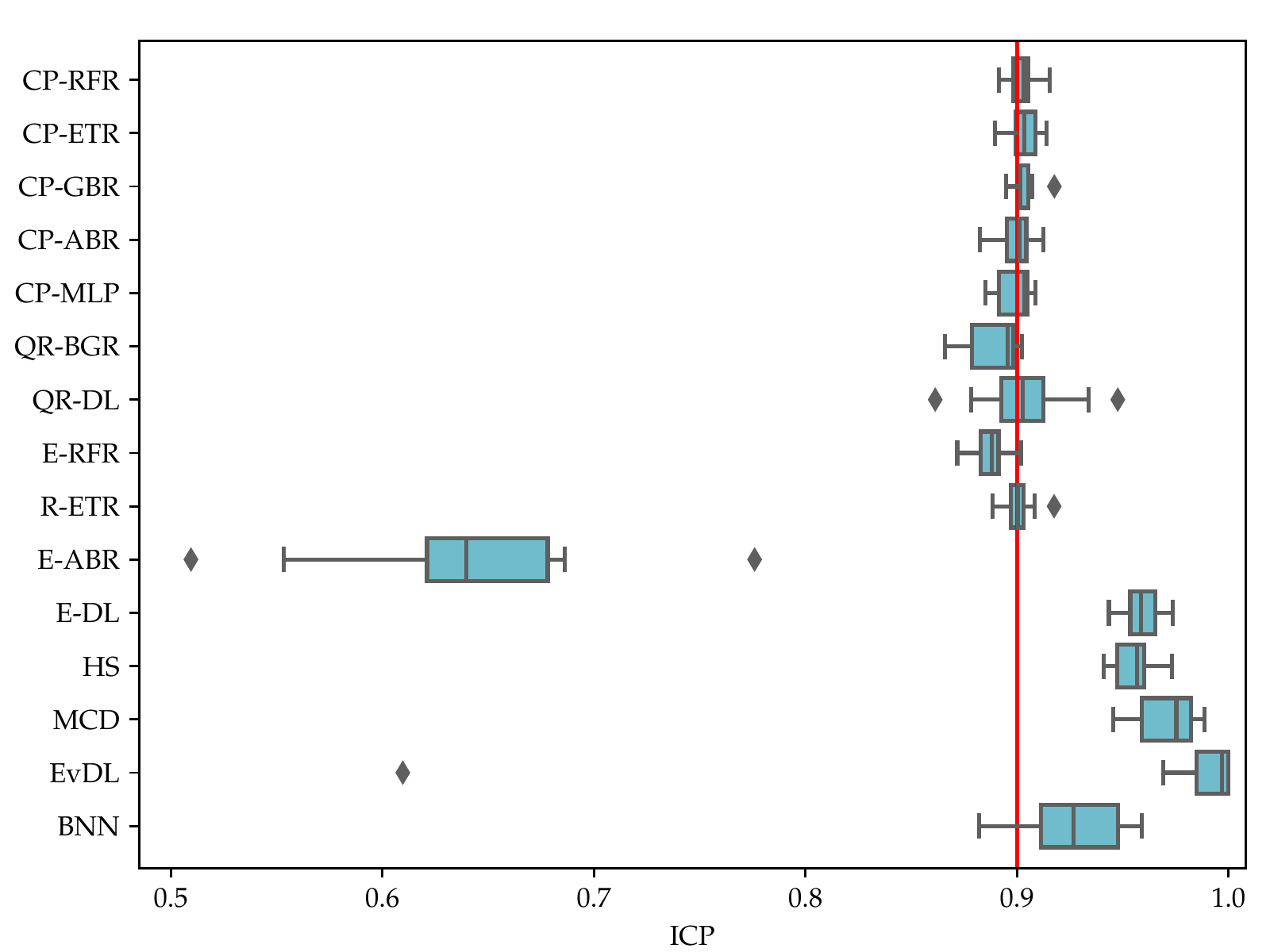}
    \caption{Interval Coverage Percentage (\texttt{ICP}) obtained by the uncertainty estimation methods over different forecasting models learned from the data collected by every loop. Boxplots represent the distribution of \texttt{ICP} values over the $10$ loops with a forecasting horizon $h=1$. Red line represents the established confidence level $\alpha=0.9$.}
    \label{fig:ICP}
\end{figure}

\begin{figure}[h!]
    \centering
    \includegraphics[width=\columnwidth]{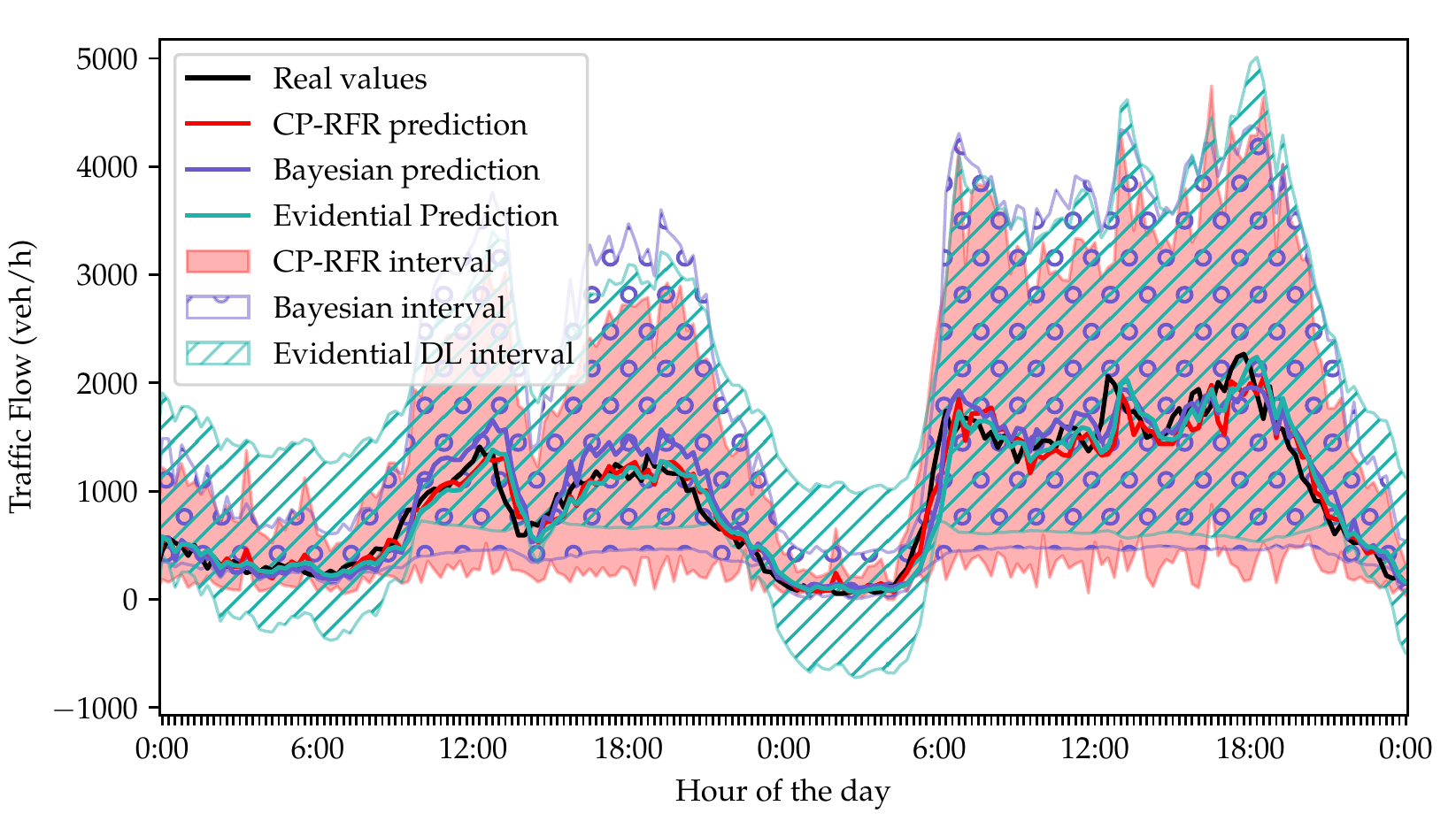}
    \caption{Two days of test traffic data collected in loop 4458. Real values (\emph{ground truth}) compared to forecasts given by three differently performing methods, together with their estimated confidence intervals for $\alpha=0.9$.}
    \label{fig:intervals_methods}
\end{figure}

Figure \ref{fig:ICP} depicts the distributions of the interval coverage \texttt{ICP} of each method over the different sensors. The red line represents the statistical significance $\alpha$ established for the confidence levels. In principle, the coverage of all methods should be close to this line, disregarding the input dataset, for the sake of a reliable behavior. It is noticeable that CP-based methods achieve a stable and close-to-$\alpha$ distribution of the \texttt{ICP} metric: this implies that the interval width values discussed before steadily cover $90\%$ of the real traffic samples, rendering them useful to know the model uncertainty. Quantile approaches have a bigger dispersion, specially for deep learning. This means that for intervals with similar width, the amount of real samples that are left out the interval is more dependent on the input data. Results for obtaining uncertainty estimates from the ensemble version of AdaBoost are clearly the worst, with confidence intervals similar in size to others, but covering much less real data (they are wider in some areas but much narrower in others, providing similar average widths, but worse coverage). This is linked to the way in which AdaBoost builds each estimator using information from the previous ones: since it is a sequential process in which every estimator specializes in difficult-to-forecast examples, weak learners composing the boosting ensemble are not by themselves reliable forecasting models \cite{schapire2013explaining}. This makes the estimation of uncertainty based on the individual predictions of such weak learners highly unreliable. Lastly, we again observe that deep learning based approaches (also HR and BNN) produce larger confidence intervals. The coverage analysis of these wider intervals reveals that, naturally, they cover more real samples. For evidential deep learning, very large intervals cover up to the entirety of the real data. This could appear as convenient, as the intervals cover more real samples, but it is far from informative and actionable, as the intervals do not comply with the established significance level: the desired output is the narrower interval that covers the established amount of samples. This can be appreciated in Figure \ref{fig:intervals_methods}, where intervals produced for two test days in loop 4458 for BNN, EvDL and CP-RFR are compared to each other. Predictive performances appear to be very similar: forecasts of BNN, in purple, show a higher deviation with respect to their ground truth at some points, yielding a slightly lower $R^2$ score. The intervals of EvDL appear to be similar for a large part of the day, but they are overestimated in the night periods, leading to less useful confidence information for actionability purposes (e.g. adaptive street lighting in the concerned part of the road network). In the case of BNN, its confidence interval seems to be more similar to that produced by CP-RFR in this particular scenario, but it tends to be wider on the upper boundary, leaving space for more samples, and thereby increasing the value of the \texttt{ICP} metric above its target confidence level $\alpha$.

On the other side, approaches featuring larger interval widths also present a higher dispersion in their measured \texttt{ICP} values, revealing their higher dependency on the input data. In general, CP-based approaches show the most stable behavior independently of data and underlying model, due to the calibration process whose importance will be analyzed in subsequent sections.

\subsection{RQ2: Which changes in the available data affect uncertainty? How can the confidence of a forecasting model be improved by changing/augmenting data at its input?}

After exploring a general performance view of all proposed methodological approaches in all available locations, the best performing model and uncertainty estimation technique is chosen in order to delve into the way in which the configuration of the input dataset $\mathcal{X}$ affects the uncertainty of forecasts issued by such models. We select the CP-RFR combination (namely, Conformal Prediction with Random Forest Regressor), so that both \texttt{ICP} and \texttt{MIL} metrics are computed over the different dataset configurations resulting from the reduction of the number of input features and the consideration of different prediction horizons $h$. 
\begin{table}[ht]
\centering
        \caption{Confidence metrics of CP-RFR over different dataset and traffic sensors}
        \vspace{3mm}
        \label{table:rq2}
        \resizebox{\columnwidth}{!}{
        \begin{tabular}{ccccccccccccccccccccc} 
         \toprule
         \multirow{1}{*}{\textbf{Sensor}}  &
         
         \multicolumn{1}{c}{Metric}  & 
        \multicolumn{4}{c}{\makecell{All Features\\($m=c=1$)}} & &
         \multicolumn{4}{c}{\makecell{Traffic + Calendar\\($m=0$, $c=1$)}} & &
         \multicolumn{4}{c}{\makecell{Traffic + Meteo\\($m=1$, $c=0$)}}& &
         \multicolumn{4}{c}{\makecell{Only Traffic\\($m=c=0$)}} \\
         \cmidrule{3-6} \cmidrule{8-11} \cmidrule{13-16} \cmidrule{18-21}
          & &
         \textit{h=1}&\textit{h=2}&\textit{h=4}&\textit{h=8} & &      \textit{h=1}&\textit{h=2}&\textit{h=4}&\textit{h=8} & &
         \textit{h=1}&\textit{h=2}&\textit{h=4}&\textit{h=8} & &
         \textit{h=1}&\textit{h=2}&\textit{h=4}&\textit{h=8}\\
         \cmidrule(lr){1-21}
\multirow{3}{*}{\textbf{4458}}
&$R^2$  &0.94 &0.92 &0.86 &0.75 &  &0.94 &0.92 &0.86 &0.75 &  &0.94 &0.91 &0.85 &0.71 &  &0.94 &0.91 &0.84 &0.69 \\ 
&\texttt{ICP}  &0.89 &0.89 &0.88 &0.88 &  &0.89 &0.89 &0.89 &0.89 &  &0.89 &0.89 &0.89 &0.89 &  &0.89 &0.89 &0.89 &0.89\\ 
&\texttt{MIL} &527.76 &640.15 &850.82 &1220.51 &  &531.66 &646.57 &865.51 &1201.72 &  &540.41 &676.08 &921.57 &1328.06 &  &557.76 &683.74 &925.55 &1292.41 \\ 
\cmidrule(lr){1-21}
\multirow{3}{*}{\textbf{6980}}
&$R^2$  &0.84 &0.81 &0.75 &0.61 &  &0.84 &0.81 &0.74 &0.61 &  &0.84 &0.8 &0.72 &0.56 &  &0.83 &0.79 &0.7 &0.53 \\ 
&\texttt{ICP}  &0.91 &0.9 &0.91 &0.9 &  &0.9 &0.9 &0.91 &0.91 &  &0.91 &0.9 &0.91 &0.91 &  &0.9 &0.9 &0.91 &0.91\\ 
&\texttt{MIL} &216.62 &231.81 &275.49 &344.23 &  &218.71 &230.96 &279.92 &352.82 &  &218.93 &241.38 &289.64 &378.44 &  &223.22 &240.9 &293.27 &380.8 \\ 
\cmidrule(lr){1-21}
\multirow{3}{*}{\textbf{10124}}
&$R^2$  &0.79 &0.77 &0.7 &0.57 &  &0.78 &0.76 &0.69 &0.56 &  &0.78 &0.76 &0.66 &0.49 &  &0.77 &0.73 &0.64 &0.45 \\ 
&\texttt{ICP}  &0.9 &0.91 &0.9 &0.89 &  &0.9 &0.9 &0.9 &0.9 &  &0.9 &0.91 &0.91 &0.9 &  &0.9 &0.9 &0.9 &0.89\\ 
&\texttt{MIL} &137.79 &146.69 &174.66 &210.66 &  &136.62 &148.33 &174.03 &214.51 &  &138.46 &154.19 &189.33 &232.42 &  &143.1 &157.88 &188.92 &232.06 \\ 
\cmidrule(lr){1-21}
\multirow{3}{*}{\textbf{6132}}
&$R^2$  &0.78 &0.76 &0.69 &0.52 &  &0.78 &0.75 &0.68 &0.5 &  &0.77 &0.74 &0.65 &0.45 &  &0.76 &0.73 &0.63 &0.42 \\ 
&\texttt{ICP}  &0.91 &0.9 &0.9 &0.9 &  &0.91 &0.9 &0.9 &0.9 &  &0.9 &0.9 &0.9 &0.9 &  &0.91 &0.91 &0.9 &0.9\\ 
&\texttt{MIL} &93.94 &97.12 &106.7 &133.95 &  &95.64 &99.38 &107.25 &133.89 &  &95.65 &99.52 &114.34 &140.92 &  &98.57 &103.42 &115.93 &147.53 \\ 
\cmidrule(lr){1-21}
\multirow{3}{*}{\textbf{3642}}
&$R^2$ &0.87 &0.84 &0.75 &0.63 &  &0.87 &0.83 &0.74 &0.63 &  &0.86 &0.82 &0.71 &0.56 &  &0.86 &0.81 &0.68 &0.53 \\ 
&\texttt{ICP}  &0.89 &0.9 &0.89 &0.9 &  &0.9 &0.9 &0.89 &0.9 &  &0.9 &0.9 &0.89 &0.91 &  &0.9 &0.9 &0.88 &0.89\\ 
&\texttt{MIL} &107.04 &121.9 &140.59 &187.3 &  &109.15 &124.02 &142.84 &185.65 &  &111.11 &126.93 &152.12 &207.48 &  &112.9 &130.03 &155.45 &196.81 \\ 
\cmidrule(lr){1-21}
\multirow{3}{*}{\textbf{4192}}
&$R^2$  &0.95 &0.93 &0.89 &0.8 &  &0.95 &0.93 &0.89 &0.81 &  &0.95 &0.93 &0.87 &0.75 &  &0.94 &0.92 &0.87 &0.75 \\ 
&\texttt{ICP}  &0.91 &0.91 &0.91 &0.91 &  &0.91 &0.91 &0.91 &0.91 &  &0.91 &0.91 &0.91 &0.91 &  &0.92 &0.91 &0.91 &0.91\\ 
&\texttt{MIL} &466.4 &541.57 &692.87 &983.08 &  &475.68 &544.31 &700.7 &967.34 &  &484.22 &565.85 &741.42 &1055.89 &  &502.96 &574.93 &757.04 &1084.64 \\ 
\cmidrule(lr){1-21}
\multirow{3}{*}{\textbf{3697}}
&$R^2$  &0.95 &0.94 &0.9 &0.81 &  &0.95 &0.94 &0.9 &0.8 &  &0.95 &0.93 &0.89 &0.79 &  &0.95 &0.93 &0.89 &0.78 \\ 
&\texttt{ICP}  &0.91 &0.9 &0.9 &0.91 &  &0.9 &0.9 &0.9 &0.9 &  &0.91 &0.9 &0.91 &0.91 &  &0.9 &0.9 &0.9 &0.92\\ 
&\texttt{MIL} &473.44 &544.55 &670.17 &959.55 &  &459.33 &547.35 &677.76 &959.58 &  &489.7 &565.44 &731.49 &1029.71 &  &485.88 &567.11 &728.05 &1055.73 \\ 
\cmidrule(lr){1-21}
\multirow{3}{*}{\textbf{3910}}
&$R^2$  &0.95 &0.93 &0.88 &0.78 &  &0.95 &0.93 &0.88 &0.79 &  &0.95 &0.92 &0.87 &0.73 &  &0.94 &0.92 &0.86 &0.72 \\ 
&\texttt{ICP}  &0.91 &0.9 &0.91 &0.9 &  &0.91 &0.9 &0.9 &0.9 &  &0.91 &0.9 &0.9 &0.9 &  &0.9 &0.9 &0.9 &0.91\\ 
&\texttt{MIL} &283.58 &331.51 &428.51 &626.99 &  &285.29 &325.0 &419.52 &595.69 &  &293.95 &347.2 &453.5 &660.41 &  &296.98 &348.3 &454.07 &657.75 \\ 
\cmidrule(lr){1-21}
\multirow{3}{*}{\textbf{3500}}
&$R^2$  &0.64 &0.62 &0.56 &0.45 &  &0.63 &0.61 &0.55 &0.47 &  &0.63 &0.6 &0.53 &0.41 &  &0.62 &0.59 &0.52 &0.41 \\ 
&\texttt{ICP}  &0.89 &0.9 &0.89 &0.9 &  &0.9 &0.9 &0.89 &0.9 &  &0.9 &0.9 &0.89 &0.9 &  &0.9 &0.89 &0.9 &0.9\\ 
&\texttt{MIL} &70.49 &74.86 &79.94 &94.59 &  &72.6 &75.89 &79.99 &93.54 &  &73.59 &76.45 &82.58 &100.66 &  &73.96 &76.03 &86.1 &99.97 \\ 
\cmidrule(lr){1-21}
\multirow{3}{*}{\textbf{5761}}
&$R^2$  &0.94 &0.92 &0.88 &0.78 &  &0.94 &0.92 &0.87 &0.77 &  &0.94 &0.91 &0.86 &0.74 &  &0.93 &0.91 &0.86 &0.73 \\ 
&\texttt{ICP}  &0.91 &0.91 &0.9 &0.9 &  &0.91 &0.9 &0.9 &0.9 &  &0.9 &0.9 &0.9 &0.9 &  &0.9 &0.9 &0.9 &0.91\\ 
&\texttt{MIL} &725.49 &843.83 &1032.46 &1369.65 &  &722.01 &835.95 &1043.94 &1417.15 &  &727.19 &849.99 &1063.37 &1441.41 &  &753.34 &854.84 &1098.85 &1474.63 \\ 
         \bottomrule
        \end{tabular}}
\end{table}

Table \ref{table:rq2} summarizes the metrics obtained in all these experiments. Several clear patterns arise in the light of these results. To begin with, an expected degradation of predictive performance occurs for all situations when the predictive horizon $h$ is increased. However, removing features from the original complete dataset has, in most cases, a slim impact in what regards to the prediction error: forecasting with all features renders very similar predictive scores that considering only traffic and calendar (i.e., disregarding meteorological features). There are cases in which removing the meteorological features produces better results (sensors 3500 and 3910) for $h=4$, from where one may conclude that these features do not contribute significantly to the model's performance in this circumstances. Analogously, when traffic and meteorological datasets are fed, the differences to the case when only traffic data is used for forecasting are negligible in most cases (with exceptions like the higher predictive horizons for sensor 10124). This minor contribution of weather features to the model's performance was identified in \cite{lana2016role}, as a consequence of an highly stable weather in this city, and the inherent relation of certain weather conditions with calendar features. This conclusion should motivate the community to go beyond predictive performance and also examine whether additional features contribute anyhow to the reduction of the uncertainty associated to the forecasts. 

Secondly, when the focus is set on the \texttt{ICP} metric, its steady behaviour throughout the whole set of scenarios is remarkable albeit expected, as the calibration stage of Conformal Prediction is precisely aimed at achieving this. The next section will revolve around this in connection to RQ3, analyzing the consequences of obtaining confidence intervals without any calibration. Furthermore, considering that the values of \texttt{ICP} guarantee that the demanded $90\%$ of true samples are inside the intervals provided in all cases, the width of such intervals can be compared fairly to each other. In general, trends noted in the \texttt{MIL} metric are similar to those found in performance measurements: the more difficult to predict, the wider the interval. However, this measurement provides another perspective of the contribution of different variables in each case. For instance, increasing the prediction horizon from $h=1$ to $h=8$ results in a $R^2$ degradation of around $0.2$ for loops 4458 and 3500. For this similar performance degradation, the confidence interval is 2.27 times wider in $h=8$ than in $h=1$ for loop 4458, while it is only 1.3 times wider in the other sensor. Thus, performance and confidence intervals do not degrade equally, because they are not co-linear. This means that performance itself is not enough to estimate uncertainty. On the other hand, it is possible to observe some situations in which removing variables favors the confidence; in some of them like in sensor 6980 with forecasting horizon $h=2$ the change is small, but the model is more confident without meteorological features: Traffic and calendar presents a narrower interval than all features, and only traffic is also narrower than traffic and meteorological features. This seems an interesting way of analyzing whether a set of features do contribute to the confidence of the model. This happens for loop 4458, where weather features have a positive impact in both cases. Moreover, there are cases in which removing a variable results in narrower intervals, such as loop 3697 with meteorological features and only traffic features. This improvement in uncertainty could be due to statistical variance of these particular data (\texttt{MIL} is computed by averaging over thousands of samples), but in any case it reinforces the intuition that these variables have a meager contribution to performance. 

When confidence estimation techniques are properly calibrated, the size of the confidence interval can be a valuable indicator of the way in which features contribute to the confidence levels of the output. Thus, besides obtaining an actionable piece of information with the confidence intervals that guarantee the inclusion of a certain amount of true samples, the analysis of these intervals allows for further insights than can be helpful when designing the dataset $\mathcal{X}$ for learning the forecasting model. In the particular datasets considered in this study, it is apparent that meteorological features contribute slightly in terms of predictive performance, but having the intervals not only confirms this point, but it even stresses it out for some sensors. For example, loop 5761 has equal results in terms of performance for all features and for traffic and calendar datasets, but the latter presents smaller confidence intervals. In conclusion: meteorological features are only contributing to the  \textit{uncertainty} of traffic forecasts.

\subsection{RQ3: What impact does the calibration process of some of the uncertainty estimation techniques have on their outcome? Why is it relevant for traffic forecasting?}

The last research question relates to the calibration of models and its impact on the quality of the estimated uncertainty. Certainly, the concept of \emph{calibration} is not new \cite{gupta2006model}, but is relatively overseen in a broad part of the literature focused on uncertainty estimation. The calibration process aims at guaranteeing the properties of the output interval. In the results shown in previous section it is possible to observe that regardless the dataset attributes, it the confidence estimation technique is calibrated, the output interval always covers a percentage of true samples very close to the significance level established by design. This means that the width of the intervals can be trusted (an important feature for the trustworthiness of the uncertainty estimation and the actionability of forecasts) and always have the same meaning among the considered combinations of models and estimation techniques. 

In order to assess the relevance of this calibration process, we note that it is an essential part of Conformal Prediction. Therefore, we perform a comparison of the aforementioned CP-RFR to a non-calibrated uncertainty estimation counterpart (E-RFR), as well as to other methods characterized by an unstable \texttt{ICP} behavior in Figure \ref{fig:ICP}, namely, HR and EvDL. The outcomes of this comparison are presented in Table \ref{table:rq3}, considering the least favorable datasets (only traffic) as the effects of the lack of calibration on the estimated uncertainty become more noticeable with less variability in the input. As the calibration process consists precisely of statistically characterizing the behavior of such a variability, in low-variability scenarios out of distribution data instances may have more weight in the definition of the intervals. Uncertainty estimation techniques that do not consider this can end up providing 1) wider intervals to cover more real samples than the specified percentage (due to the fact that they consider for the intervals points that are outside the significance boundaries in the training data distribution); or 2) narrower intervals that do not meet the statistical significance specifications.
\begin{table}[ht]
\centering
        \caption{Confidence metrics of calibrated and non-calibrated approaches over scenarios with only traffic data ($m=c=0$).}
        \label{table:rq3}
        \vspace{3mm}
        \resizebox{\columnwidth}{!}{
        \begin{tabular}{ccccccccccccccccccccc} 
         \toprule
         \multirow{1}{*}{\textbf{Sensor}}  &
         
         \multicolumn{1}{c}{Metric}  & 
        \multicolumn{4}{c}{CP-RFR} & &
         \multicolumn{4}{c}{E-RFR} & &
         \multicolumn{4}{c}{HR}& &
         \multicolumn{4}{c}{EvDL} \\
         \cmidrule{3-6} \cmidrule{8-11} \cmidrule{13-16} \cmidrule{18-21}
          & &
         \textit{h=1}&\textit{h=2}&\textit{h=4}&\textit{h=8} & &      \textit{h=1}&\textit{h=2}&\textit{h=4}&\textit{h=8} & &
         \textit{h=1}&\textit{h=2}&\textit{h=4}&\textit{h=8} & &
         \textit{h=1}&\textit{h=2}&\textit{h=4}&\textit{h=8}\\
         \cmidrule(lr){1-21}
\multirow{3}{*}{\textbf{4458}}
&$R^2$  &0.94 &0.91 &0.84 &0.69 &  &0.94 &0.91 &0.84 &0.69 &  &0.94 &0.91 &0.83 &0.64 &  &0.91 &0.87 &0.77 &0.47 \\ 
&\texttt{ICP}  &0.89 &0.89 &0.89 &0.89 &  &0.84 &0.85 &0.85 &0.83 &  &0.98 &0.97 &0.98 &0.99 &  &0.98 &0.93 &0.74 &0.23\\ 
&\texttt{MIL} &557.76 &683.74 &925.55 &1292.41 &  &450.69 &545.88 &735.34 &1015.81 &  &709.93 &874.9 &1239.66 &2030.38 &  &1462.95 &940.43 &573.01 &180.88 \\ 
\cmidrule(lr){1-21}
\multirow{3}{*}{\textbf{6980}}
&$R^2$  &0.83 &0.79 &0.7 &0.53 &  &0.83 &0.79 &0.7 &0.53 &  &0.83 &0.79 &0.7 &0.53 &  &0.79 &0.72 &0.56 &0.31 \\ 
&\texttt{ICP}  &0.9 &0.9 &0.91 &0.91 &  &0.86 &0.87 &0.87 &0.87 &  &0.99 &0.98 &0.98 &0.99 &  &1.0 &1.0 &1.0 &0.99\\ 
&\texttt{MIL} &223.22 &240.9 &293.27 &380.8 &  &179.96 &202.18 &243.01 &301.56 &  &292.54 &332.53 &403.14 &520.31 &  &1564.44 &1272.89 &948.28 &915.4 \\ 
\cmidrule(lr){1-21}
\multirow{3}{*}{\textbf{10124}}
&$R^2$  &0.77 &0.73 &0.64 &0.45 &  &0.77 &0.73 &0.64 &0.46 &  &0.78 &0.73 &0.62 &0.45 &  &0.74 &0.68 &0.53 &0.27 \\ 
&\texttt{ICP}  &0.9 &0.9 &0.9 &0.89 &  &0.86 &0.86 &0.86 &0.86 &  &0.96 &0.96 &0.96 &0.97 &  &1.0 &0.99 &0.96 &0.81\\ 
&\texttt{MIL} &143.1 &157.88 &188.92 &232.06 &  &115.38 &126.85 &146.69 &176.21 &  &155.12 &176.96 &218.54 &300.66 &  &488.73 &378.8 &312.39 &165.98 \\ 
\cmidrule(lr){1-21}
\multirow{3}{*}{\textbf{6132}}
&$R^2$  &0.76 &0.73 &0.63 &0.42 &  &0.76 &0.73 &0.63 &0.42 &  &0.77 &0.74 &0.64 &0.45 &  &0.71 &0.67 &0.49 &0.33 \\ 
&\texttt{ICP}  &0.91 &0.91 &0.9 &0.9 &  &0.86 &0.86 &0.87 &0.87 &  &0.97 &0.97 &0.97 &0.97 &  &0.99 &0.98 &0.89 &0.49\\ 
&\texttt{MIL} &98.57 &103.42 &115.93 &147.53 &  &76.42 &80.95 &92.79 &113.56 &  &103.83 &112.81 &134.65 &183.4 &  &164.28 &155.25 &110.06 &50.95 \\ 
\cmidrule(lr){1-21}
\multirow{3}{*}{\textbf{3642}}
&$R^2$  &0.86 &0.81 &0.68 &0.53 &  &0.85 &0.81 &0.68 &0.53 &  &0.84 &0.78 &0.66 &0.5 &  &0.83 &0.75 &0.59 &0.32 \\ 
&\texttt{ICP}  &0.9 &0.9 &0.88 &0.89 &  &0.86 &0.86 &0.86 &0.85 &  &0.97 &0.97 &0.98 &0.98 &  &0.98 &0.97 &0.82 &0.55\\ 
&\texttt{MIL} &112.9 &130.03 &155.45 &196.81 &  &90.19 &103.11 &127.87 &157.05 &  &134.26 &162.31 &216.65 &291.06 &  &218.23 &208.17 &120.27 &76.29 \\ 
\cmidrule(lr){1-21}
\multirow{3}{*}{\textbf{4192}}
&$R^2$  &0.94 &0.92 &0.87 &0.75 &  &0.94 &0.92 &0.87 &0.75 &  &0.94 &0.91 &0.85 &0.71 &  &0.92 &0.89 &0.78 &0.61 \\ 
&\texttt{ICP}  &0.92 &0.91 &0.91 &0.91 &  &0.85 &0.85 &0.85 &0.85 &  &0.96 &0.97 &0.97 &0.98 &  &1.0 &0.94 &0.58 &0.35\\ 
&\texttt{MIL} &502.96 &574.93 &757.04 &1084.64 &  &375.39 &441.29 &569.86 &775.43 &  &585.99 &727.23 &995.07 &1460.43 &  &2831.65 &772.97 &357.99 &298.92 \\ 
\cmidrule(lr){1-21}
\multirow{3}{*}{\textbf{3697}}
&$R^2$  &0.95 &0.93 &0.89 &0.78 &  &0.95 &0.93 &0.89 &0.78 &  &0.95 &0.93 &0.87 &0.75 &  &0.93 &0.89 &0.77 &0.59 \\ 
&\texttt{ICP}  &0.9 &0.9 &0.9 &0.92 &  &0.85 &0.85 &0.85 &0.85 &  &0.98 &0.98 &0.98 &0.97 &  &0.95 &0.89 &0.4 &0.2\\ 
&\texttt{MIL} &485.88 &567.11 &728.05 &1055.73 &  &380.29 &447.0 &569.21 &789.45 &  &646.82 &765.92 &992.71 &1391.19 &  &720.45 &686.63 &228.62 &142.76 \\ 
\cmidrule(lr){1-21}
\multirow{3}{*}{\textbf{3910}}
&$R^2$  &0.94 &0.92 &0.86 &0.72 &  &0.94 &0.92 &0.86 &0.72 &  &0.94 &0.91 &0.84 &0.67 &  &0.91 &0.86 &0.63 &0.38 \\ 
&\texttt{ICP}  &0.9 &0.9 &0.9 &0.91 &  &0.86 &0.85 &0.86 &0.86 &  &0.97 &0.97 &0.97 &0.97 &  &0.99 &0.97 &0.53 &0.67\\ 
&\texttt{MIL} &296.98 &348.3 &454.07 &657.75 &  &232.68 &276.74 &362.88 &505.82 &  &345.72 &424.19 &716.0 &961.62 &  &852.71 &726.62 &192.17 &368.8 \\ 
\cmidrule(lr){1-21}
\multirow{3}{*}{\textbf{3500}}
&$R^2$  &0.62 &0.59 &0.52 &0.41 &  &0.62 &0.59 &0.52 &0.41 &  &0.65 &0.62 &0.56 &0.44 &  &0.61 &0.6 &0.52 &0.38 \\ 
&\texttt{ICP}  &0.9 &0.89 &0.9 &0.9 &  &0.86 &0.86 &0.86 &0.86 &  &0.97 &0.97 &0.97 &0.97 &  &1.0 &1.0 &1.0 &1.0\\ 
&\texttt{MIL} &73.96 &76.03 &86.1 &99.97 &  &60.47 &64.93 &72.12 &81.35 &  &96.12 &103.35 &115.55 &135.03 &  &880.21 &927.87 &769.13 &530.5 \\ 
\cmidrule(lr){1-21}
\multirow{3}{*}{\textbf{5761}}
&$R^2$  &0.93 &0.91 &0.86 &0.73 &  &0.93 &0.91 &0.86 &0.73 &  &0.93 &0.91 &0.85 &0.7 &  &0.91 &0.89 &0.76 &0.61 \\ 
&\texttt{ICP} &0.9 &0.9 &0.9 &0.91 &  &0.85 &0.85 &0.85 &0.85 &  &0.97 &0.97 &0.97 &0.98 &  &0.55 &0.71 &0.18 &0.41\\ 
&\texttt{MIL} &753.34 &854.84 &1098.85 &1474.63 &  &581.63 &675.8 &855.98 &1151.84 &  &895.51 &1044.98 &1434.81 &2144.94 &  &357.83 &551.66 &132.8 &480.21 \\ 
         \bottomrule
        \end{tabular}}
\end{table}

The first 4 columns of Table \ref{table:rq3} correspond to the calibrated method: in them, it is possible to observe the persistence of the \texttt{ICP} value for all loops and all values of the forecasting horizon $h$. When considering the same forecasting model (RFR) but using the ensemble technique to estimate the intervals, very similar $R^2$ scores are obtained (as expected, as they essentially resort to the same algorithm for producing the forecasts). However, in this case the whole set of results appears to be under-calibrated, as the coverage of the estimated intervals is lower than the expected one (85\%), giving place to narrower intervals. Should these intervals and performance levels be considered to compare them to those elicited by CP-RFR, the comparison would declare E-RFR as the best forecasting method. Nonetheless, it is leaving 5\% of the real samples outside the coverage of its confidence intervals, thereby penalizing the trustworthiness of the estimated uncertainty. HR seems to work the other way around, by increasing the size of the interval up to covering almost all real cases (\texttt{ICP} close to 99\%), and obtaining similar regression results. Lastly, EvDL was shown in the previous section that it failed to provide reliable confidence levels with the complete dataset and horizon $h=1$: the reported interval width tended to be wider and \texttt{ICP} values were close to 1 in most cases. When only-traffic data were provided, this method performed erratically. Apparently, wider intervals and \texttt{ICP} values close to 1 are consistent with $h=1$, but while for some sensors this behavior is sustained, others present a notable decay in predictive performance (see the cases of sensors 4458, 6980, 3910, in comparison to other methods), and also in the size of the interval, reducing the coverage to very low levels (even to only 20\% of samples in loop 3697). This demeanor is sharper for those sensors with larger dynamic ranges, suggesting the susceptibility of the method to deal with data in the tails of the distribution unless 1) its regularization parameter $\lambda$ is properly tuned; or 2) by performing a posterior calibration of the model's output.

\begin{table}[ht]
\centering
\vspace{-3mm}
        \caption{Confidence metrics for CP-RFR with different confidence levels $\alpha$ compared to other methods with a fixed confidence level.}
        \label{table:rq32}
        \vspace{3mm}
        \resizebox{0.6\columnwidth}{!}{\begin{tabular}{cccccc} 
         \toprule
         \multirow{2}{*}{\makecell{\textbf{Sensor}\\\textbf{5761}}}  &
         & \multicolumn{4}{c}{Metric}  \\
         \cmidrule{3-6} 
          & &
         \textit{h=1}&\textit{h=2}&\textit{h=4}&\textit{h=8} \\
         \cmidrule(lr){1-6}
           \multirow{2}{*}{\makecell{CP-RFR\\$\alpha=0.85$}} &\texttt{ICP}  &0.86 &0.85 &0.85 &0.85 \\ 
            &\texttt{MIL} &627.12 &716.45 &918.45 &1214.01 \\ 
            \midrule
            \multirow{2}{*}{\makecell{E-RFR\\$\alpha=0.9$}}& \texttt{ICP}  & 0.85 &0.85 &0.85 &0.85 \\ 
            & \texttt{MIL} &581.63 &675.8 &855.98 &1151.84 \\ 
            \midrule
            \multirow{2}{*}{\makecell{CP-RFR\\$\alpha=0.99$}} & \texttt{ICP} &0.99 &0.99 &0.99 &0.99 \\ 
            & \texttt{MIL} &1535.12 &1896.72 &2596.84 &3668.75 \\ 
            \midrule
            \multirow{2}{*}{\makecell{HR\\$\alpha=0.9$}} &\texttt{ICP} &0.97 &0.97 &0.97 &0.98\\ 
            &\texttt{MIL} &895.51 &1044.98 &1434.81 &2144.94 \\ 
         \bottomrule
        \end{tabular}}
\end{table}
 
Additional tests were conducted considering CP-RFR and levels of confidence equal to $\alpha=0.85$ and $\alpha=0.99$ in order to assess if such methods are just providing the intervals that are proper for other levels of confidence. Table \ref{table:rq32} reports, for one of the sensors, a trend that prevails over the totality of loops analyzed in these experiments: in general, intervals obtained for $\alpha=0.9$ that cover 85\% of the real traffic samples are narrower than confidence intervals provided by CP-RFR for $\alpha=0.85$. The same behavior occurs when $\alpha=0.99$. One may arrive at the conclusion that intervals computed by these methods for $\alpha=0.9$ are better to estimate intervals with support equal to 0.85 and 0.99, as they provide narrower results; however, and unlike with CP, these results lack any statistically guarantees. 

\begin{figure}[h!]
    \centering
    \includegraphics[width=\columnwidth]{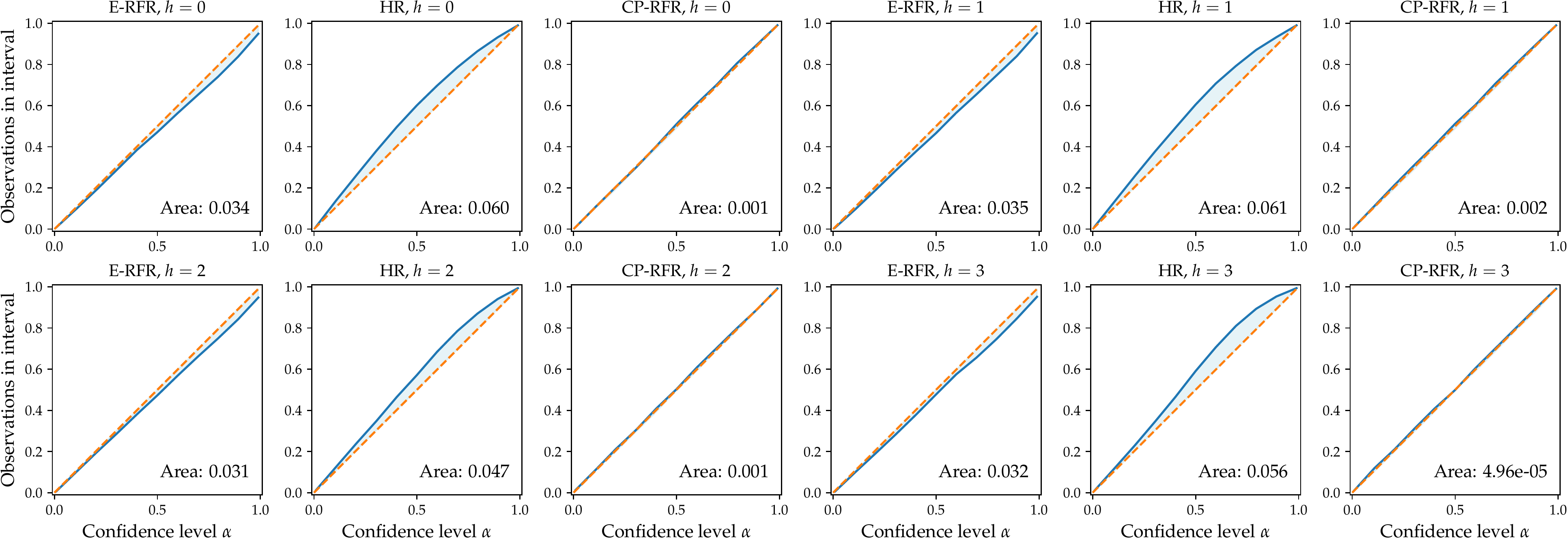}
    \caption{Calibration curves (colored in blue) corresponding to each case in Table \ref{table:rq32}. The dashed orange line corresponds to the ideal case where the confidence intervals estimated for the model are perfectly calibrated. The area between the calibration curve and the ideal case is given in each figure.}
    \label{fig:cal}
\end{figure}

To further argue on the need for calibrated traffic forecasts, we inspect the relationship between different confidence levels $\alpha$ and the observed traffic values for the forecasting horizon values and uncertainty estimation techniques reported in Table \ref{table:rq32}. Figure \ref{fig:cal} shows the calibration curves (also referred to as \emph{reliability diagrams} for each of these cases. A calibration plot examines whether the confidence interval estimated for a confidence level $\alpha$ actually captures a fraction $\alpha$ of the observed test values. The area between the curve obtained for different $\alpha$ values and the ideal case (a perfectly calibrated model) is a numerical indicator of how miscalibrated the model can be regarded to be. Plots included in this figure reveal that both E-RFR and HR are not properly calibrated for high values of $\alpha$: E-RFR is slightly overconfident in its estimated uncertainty (i.e., the computed intervals are too narrow for what they should be), whereas HR is found to be underconfident in this same region of $\alpha$ values (namely, its estimated intervals cover are too wide). By contrast, the calibration process performed in conformal prediction allows computing accurate interval estimates for any value of $\alpha$, as exposed by the closeness of its calibration curves to the ideal case and the notably smaller miscalibration area annotated inside the plots.

\section{Takeaway Messages and Outlook}\label{sec:conc}

As in many other areas related to ITS, traffic forecasting has widely embraced the irruption of data-based modeling approaches relying on Machine Learning algorithms. Advances held over the years have achieved in a pursuit towards solutions capable of producing forecasts of ever-growing precision, exploiting efficiently relationships held within traffic data flows over space and time. Lately, performance-driven research studies are progressively steering towards the trustworthiness and actionability of traffic forecasts produced by such data-based models, in view of the narrow performance gaps attained by modern modeling choices.

In this context, this paper has aimed at bringing the attention of the community working in traffic forecasting to this matter. Among the manifold perspectives from which one can favor the trustworthiness of data-based models, we have emphasized on the need for quantifying the confidence of the data-based model associated to its predicted traffic values. Assessing the uncertainty propagated to the model's output allows a traffic manager to better design countermeasures against future congestion events in the road network, delineate better traffic light schedules, decide where to deploy new traffic sensors and collect data therefrom, or quantify whether forecasts can be predicted more confidently if the model is supplied with data supplied by new sources of information. In short, uncertainty quantification (also referred to as model's confidence) is a key for human decision making based on the output of traffic forecasting models. Besides providing this rationale, this work provides an overview on the most representative uncertainty estimation methods, as well as quantitative measures to gauge the quality of the estimated confidence intervals. This first part of the manuscript pretends to be a soft entry point to newcomers interested in confidence-aware traffic forecasting, establishing the main motivational reasons for research in the area, essential information pointers, a summary of baseline techniques and a description of evaluation measures and protocols.

To further complement this material, a comprehensive set of experiments over real traffic data collected in the city of Madrid (Spain) has been designed, comprising different sets of features, uncertainty estimation techniques and Machine Learning based forecasting models. Results stemming from this setup have lead to several lessons learned about the role of confidence/uncertainty in the actionability of traffic forecasts. In the first place, a diversity of uncertainty estimation techniques has been compared to each other, evincing that there is not a single source of uncertainty, nor is there a unique way of approaching its calculation. Nevertheless, the metrics used to assess the validity and informational efficiency of the confidence intervals estimated by such techniques has uncovered that Conformal Prediction is more stable and reliable than approaches based on the ablation of models to statistically characterize their output (e.g. ensemble models or Monte Carlo dropout), but also better than models specifically suited for the purpose (e.g., Bayesian Neural Networks). Besides, Conformal Prediction yields a transparent and traceable way of obtaining confidence intervals, and the calibration stage that lies at the core of its procedure helps maintain the size of the intervals to the minimum that guarantees the coverage of unseen test samples. This feature, combined with its model-agnostic nature, renders Conformal Prediction as a very interesting option for estimating the uncertainty of traffic forecasts. On the other side, methods like evidential Deep Learning fail to produce reliable intervals due to its known susceptibility to the value of its parameters and the suitability of the evidential priors. Other methods yield different levels of quality in regards to their estimated confidence intervals, which depend on the characteristics present in the input data from which traffic forecasts are predicted.Our experiments have also confirmed that calibration is essential to reliably estimate the uncertainty of forecasts for a given confidence level.

Further along this line, we have verified that the particularities of the dataset at hand may affect the amount of uncertainty associated to the model's output. In the traffic dataset used for the experiments, the stability and inherent predictive potential of the traffic time series cause that other sources of data that could potentially help to deliver more precise forecasts have a low relevance and a low impact in uncertainty. In general, short-term traffic forecasting models can benefit marginally from additional sources of data, due to the acknowledged relevance of the time series variables that take part in the model. However, there are cases in which the impact becomes noticeable in the precision of traffic forecasts, whereas in other cases the addition of new data sources produces similar performance results, yet worst confidence intervals. These cases prove that the composition of the datasets used for traffic forecasting can also be driven by the examination of its consequences for the uncertainty of the model. Therefore, a principled assessment of the uncertainty of traffic forecasts throughout the modeling pipeline and the actionability that it grants should be of pivotal importance in future studies, considering it as an additional dimension in comparison benchmarks, and as a criterion to decide whether to include new sources of information (e.g. social media, traffic cameras) that may jeopardize the confidence of the model in its predictions.

Beyond the questions addressed experimentally in this research work, other applications of uncertainty estimation that were presented in Section \ref{sec:uncertainty} have not been explored to date. We envision a rich agenda, plenty of uncharted research lines, related to confidence-aware traffic forecasting. Among them, we remark the high stability of properly calibrated intervals, which can be helpful for trend change detection over traffic data streams. Indeed, a continuously flowing stream of traffic data should fall within the estimated confidence interval, since it statistically represents the \emph{regular} behavior of traffic data in the location of interest. If real samples eventually start falling out the confidence interval, this can be symptomatic of a contextual change of the traffic behavior (due to e.g., roadworks or any exceptional circumstance like an accident), so that a closer inspection can be enforced or an update of the traffic forecasting model be triggered upon its occurrence. Lastly, a promising application of uncertainty estimation techniques arises from its natural connection to the active learning research area: by measuring confidence intervals over different regions of the feature space at the input of the model, one can identify where most of its uncertainty is concentrated, contributing to the identification of suitable locations in the road network and/or periods in time where more traffic data should be collected. This information can be ultimately be used for designing new traffic measurement campaigns with provisionally deployed sensors over the city, showcasing another purpose for which uncertainty estimations can effectively contribute to the actionability of traffic forecasts.

\section*{Acknowledgments}

This work has received funding support from the European Union’s Horizon 2020 program (project \emph{URBANITE: Supporting the decision-making in urban transformation with the use of disruptive technologies}, grant agreement 870338), as well as from the Basque Government (ELKARTEK program and the Consolidated Research Group MATHMODE, ref. IT1256-22). 





\bibliography{bibliograf}
\bibliographystyle{IEEEtran}
\end{document}